\documentclass[11pt]{idea_style}
\newif\ifdraft
\draftfalse

\usepackage[T1]{fontenc}
\usepackage{url}
\usepackage{nicefrac}
\usepackage{multicol}
\usepackage{microtype}
\usepackage{geometry}
\usepackage{graphicx}%
\graphicspath{{fig/}}
\usepackage{multirow}%
\usepackage{amsmath}
\usepackage{amssymb}
\usepackage{amsfonts}
\usepackage{stmaryrd}
\usepackage{mathrsfs}%
\usepackage{xcolor}%
\usepackage{textcomp}%
\usepackage{manyfoot}%
\usepackage{booktabs}%
\usepackage{algorithm}%
\usepackage{tabularx}%
\usepackage{algorithmicx}%
\usepackage{algpseudocode}%
\usepackage{listings}%
\usepackage[inline]{enumitem}
\usepackage{xspace}
\usepackage{tikz}
\usepackage{makecell}
\usepackage{longtable}
\usepackage{changepage}
\usepackage{color}
\usepackage{colortbl}
\usepackage{pifont}
\usepackage{subcaption}
\usepackage{physics}
\usepackage{diagbox}
\usepackage{wrapfig}
\usepackage{siunitx}
\usepackage[misc]{ifsym} 
\usepackage[numbers]{natbib}

\usepackage{hyperref}
\usepackage{url}
\usepackage{multicol} 
\usepackage{array} 
\usepackage{threeparttable}
\usepackage{cutwin}
\usepackage{subcaption}
\usepackage{graphicx} 
\usepackage{wrapfig}
\usepackage{xcolor}
\usepackage{colortbl}
\usepackage{booktabs}
\usepackage{multirow}
\usepackage{dsfont}
\usepackage{pifont}
\usepackage{float}
\definecolor{citecolor}{HTML}{114083}
\definecolor{linkcolor}{HTML}{ED1C24}
\hypersetup{colorlinks=true,citecolor=citecolor, linkcolor=linkcolor, urlcolor=linkcolor}
\usepackage{booktabs} 

\usepackage{hyperref}
\hypersetup{
  colorlinks,
  linkcolor=ideacolor,
  citecolor=ideacolor,
  urlcolor=ideacolor
}

\usepackage{setspace}
\usepackage{changepage}
\usepackage{float}
\usepackage{tablefootnote}
\usepackage{threeparttable}
\usepackage{placeins}

\usepackage{amsmath,amsfonts,bm}

\def\eqref#1{equation~\ref{#1}}

\def\1{\bm{1}}

\DeclareMathAlphabet{\mathsfit}{\encodingdefault}{\sfdefault}{m}{sl}
\SetMathAlphabet{\mathsfit}{bold}{\encodingdefault}{\sfdefault}{bx}{n}

\definecolor{myorange}{RGB}{220,128,0}
\definecolor{myblue}{RGB}{0,80,220}
\definecolor{myred}{RGB}{220,20,0}

\newcommand{\ie}{\emph{i.e.},\xspace}
\newcommand{\eg}{\emph{e.g.},\xspace}

\def\ourname{PEAR}

\ifdraft
    \providecommand\todo[1]{[\textcolor{red}{TODO: {#1}}]}
\else
    \providecommand\todo[1]{}
\fi

\author[]{Jiahao Wu, Yunfei Liu$^{(\textrm{\Letter})}$, Lijian Lin, Ye Zhu, Lei Zhu, Jingyi Li, Yu Li}

\affiliation{International Digital Economy Academy}

\ideadata[$^{(\textrm{\Letter})}$ ]{Corresponding author.}
\ideadata[Project page]{\url{https://wujh2001.github.io/PEAR}}

\title{\includegraphics[height=1.2em]{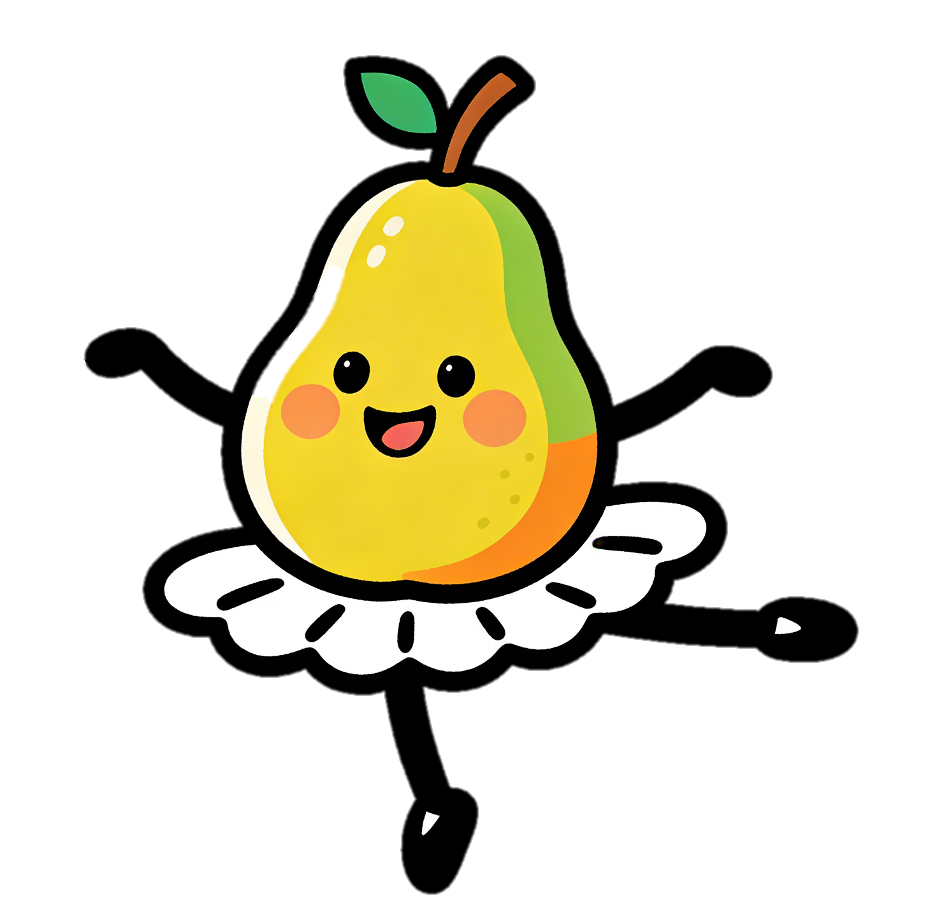}  {PEAR:}\\[7pt]\Large Pixel-Aligned Expressive Human Mesh Recovery}

\abstract{
\small{

Reconstructing detailed 3D human meshes from a single in-the-wild image remains a fundamental challenge in computer vision. Existing SMPLX-based methods often suffer from slow inference, produce only coarse body poses, and exhibit misalignments or unnatural artifacts in fine-grained regions such as the face and hands. These issues make current approaches difficult to apply to downstream tasks.
To address these challenges, we propose \ourname—a fast and robust framework for pixel-aligned expressive human mesh recovery. \ourname~ explicitly tackles three major limitations of existing methods: slow inference, inaccurate localization of fine-grained human pose details, and insufficient facial expression capture.
Specifically, to enable real-time SMPLX parameter inference, we depart from prior designs that rely on high-resolution inputs or multi-branch architectures. Instead, we adopt a clean and unified ViT-based model capable of recovering coarse 3D human geometry. To compensate for the loss of fine-grained details caused by this simplified architecture, we introduce pixel-level supervision to optimize the geometry, significantly improving the reconstruction accuracy of fine-grained human details. To make this approach practical, we further propose a modular data annotation strategy that enriches the training data and enhances the robustness of the model. Overall, \ourname ~is a preprocessing-free framework that can simultaneously infer EHM-s (SMPLX and scaled-FLAME) parameters  at over 100 FPS. Extensive experiments on multiple benchmark datasets demonstrate that our method achieves substantial improvements in pose estimation accuracy compared to previous SMPLX–based approaches.

}
}

\begin{document}

\maketitle

\begin{figure}[h]
    \centering
    \includegraphics[width=\textwidth, trim=0cm 2cm 0cm 0cm clip]{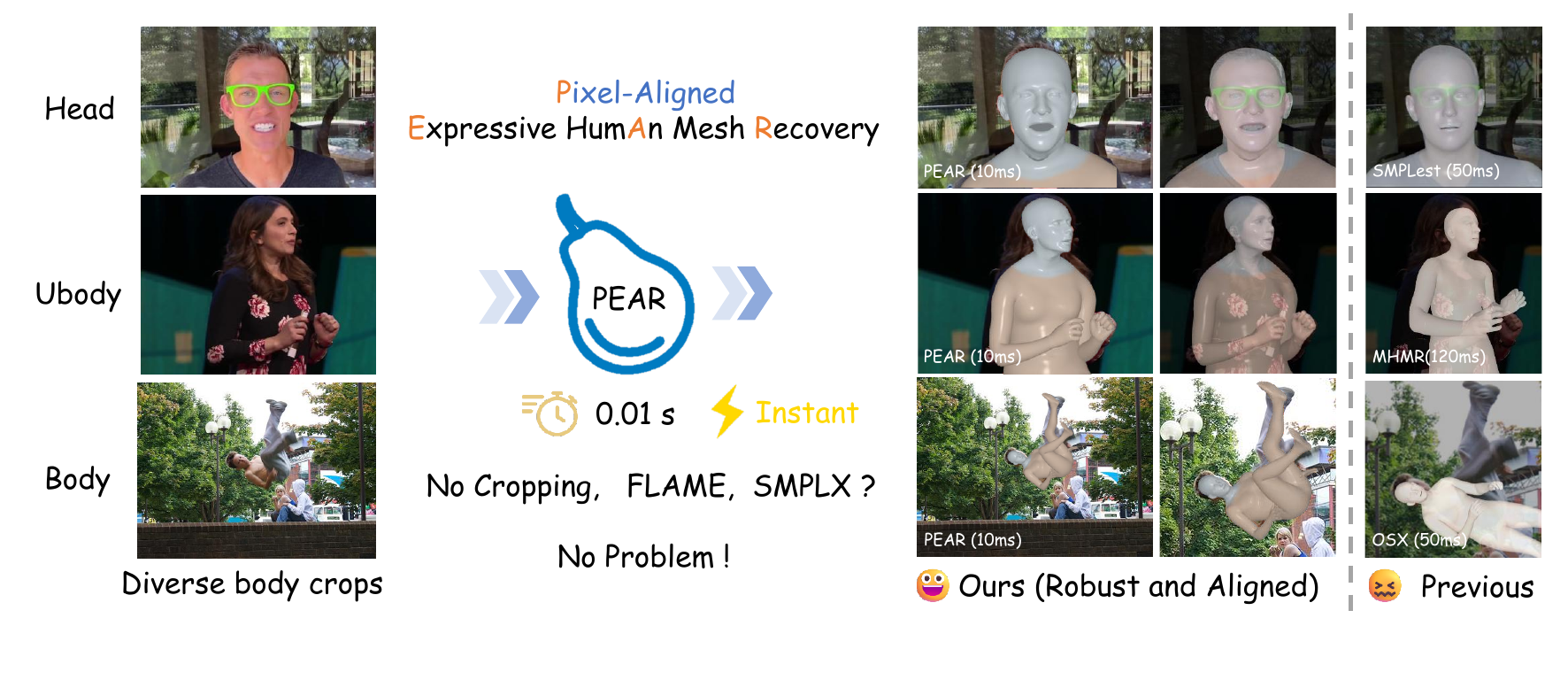}
    \caption{ We propose \textbf{PEAR}, a pixel-aligned human mesh recovery framework that surpasses  prior SMPLX-based methods and is robust to diverse human body crops. It recovers more accurate facial details and body poses from a single image in under 0.01s without body-part cropping, making it well-suited for realtime downstream applications. }
    \label{fig:teaser}
\end{figure}

\section{Introduction}

3D human pose estimation and Human Mesh Recovery (HMR) constitutes a cornerstone of computer vision, with wide-ranging applications spanning robotic perception \cite{fu2024humanplus,li2024okami}, immersive gaming, embodied AI, and the creation of digital avatars for film and telepresence \cite{rauschnabel2022xr, burdea2003virtual, speicher2019mixed,jin2023capture}. Recent advances in this domain have been propelled by parametric human models, notable SMPL \cite{SMPL:2015}, SMPLX \cite{SMPL-X:2019}, and GHUM \cite{xu2020ghum}. These models offer compact, low-dimensional manifolds that map high-dimensional human geometry, enabling neural networks to regress body parameters directly for monocular imagery and significantly advancing the state-of-the-art.

\begin{figure}[t]
  \includegraphics[width=\linewidth]{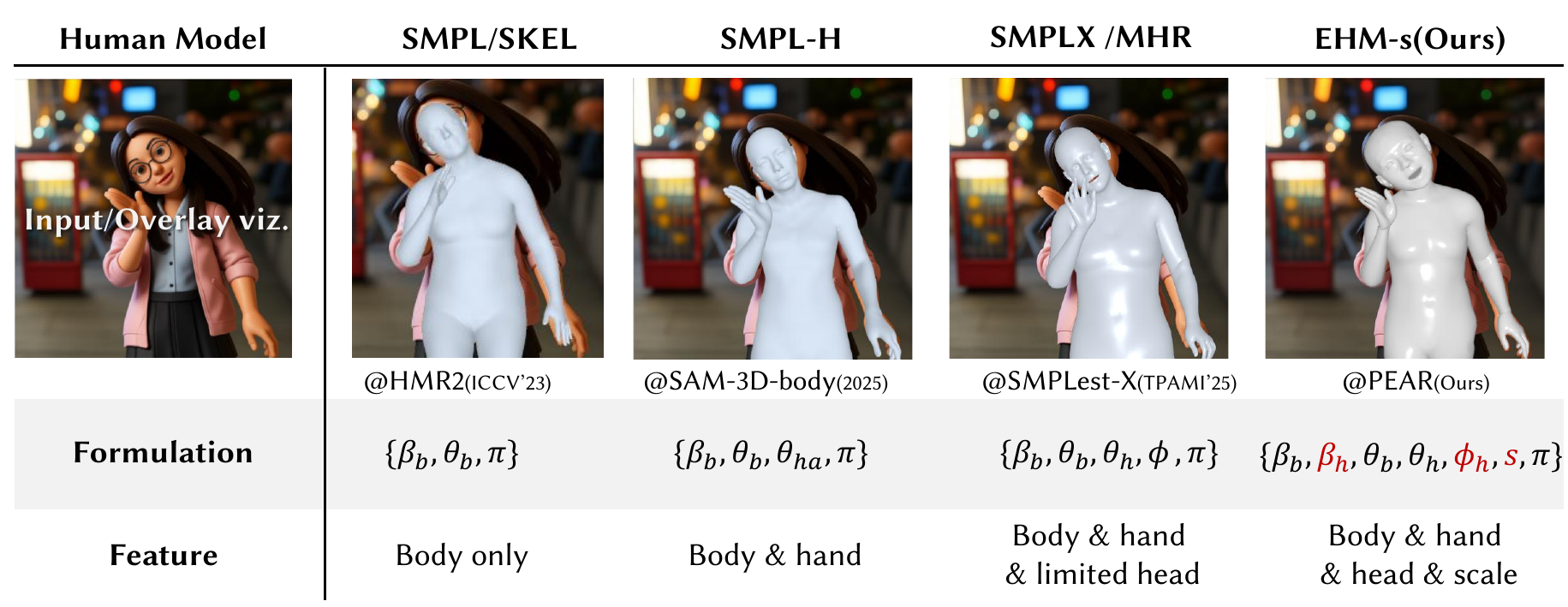}
  \caption{Current methods overview. MHR is a full-body human model; however, SAM3D-body predicts only the body and hand components of MHR. }
  \label{fig: methods comp}
\end{figure}

Existing approaches can be broadly categorized into body-only regressors (\eg, SMPL-based methods\cite{keller2023skel}) and expressive whole-body methods (\eg, MHR\cite{MHR:2025},SMPLX\cite{yin2025smplest}). \textbf{For downstream tasks requiring high-fidelity interactions, such as interpreting facial expressions or intricate hand gestures, SMPLX is a suitable choice.} However, despite its capabilities, current SMPLX-based methods \cite{baradel2024multi,lin2023one,yin2025smplest,zhang2023pymaf} still face three key challenges:
\textit{(1) Limited expressiveness.} The standard SMPLX shape space entangles cranial features with body shapes, limiting the ability to model diverse head-to-body proportions and fine-grained expressions.
\textit{(2) Pixel-level misalignment.} Reliance on sparse keypoint supervision often results in sliding artifacts and misalignment in fine-grained regions (face and hands).
\textit{(3) Computational bottlenecks.} Complex multi-branch architectures or multi-stage designed to mitigate the above issue often prevent real-time deployment and downstream applications.

To elaborate, while SMPLX integrates facial topology, it offers limited degrees of freedom or and geometry compared to specialized face models (\ie, FLAME). This constraint hampers the capture of subtle identity cues and interactive semantics (see Fig.~\ref{fig: methods comp}). 
Inspired by the Expressive Human Model (EHM)~\cite{zhang2025guava}, which integrates the FLAME head to the SMPLX body for optimization tasks, we propose a learnable, regression-friendly variant named \textbf{EHM-s}. Crucially, we introduce a global head-scale parameter $s$. Such design decouples head size from body shape, enabling our model to generalize to diverse anthropometric distributions (\eg, toddlers or stylized characters), where head proportions deviate from the adult norm.

Furthermore, regarding \textbf{alignment and speed}, prior methods  often prioritize global structural consistency at the expense of local pixel-level alignment. While optimization-based refinement \cite{moon2024exavatar,zhang2025guava} can alleviate misalignment, it is prohibitively rather slow. Conversely, regression-based methods\cite{baradel2024multi,yin2025smplest} typically resort to high-resolution inputs or multi-branch designs (\eg, separate encoders for face/hand/body) to preserve detail, which increases computations overhead and slow-down inference speed. We argue that a streamlined architecture can achieve both high fidelity and speed if supervised effectively.

To this end, we present \ourname, the first framework capable of regression expressive EHM-s parameters at \textbf{over 100 FPS}. Departing from complex multi-branch designs, we employ a single, lightweight Vision Transformer (ViT) as the unified backbone. To overcome the accuracy limitations inherent in simple architectures, we introduce a novel \textbf{analysis-by-synthesis training paradigm} for HMR. Specifically, we integrate a differentiable neural renderer that enforces dense pixel-level alignment between the predicted mesh and the input image. This allows the model to learn fine-grained details from raw pixels without incurring inference-time costs. Additionally, we redesign the data pipeline by decomposing the human mesh into modular components (body, face, hands). This modularity allows out model to learn robustly from partial inputs, handling head-only, upper-body, and whole body images seamlessly (see Fig~\ref{fig:teaser}).

In summary, our contributions are: 
\begin{itemize}
    \item   We propose  PEAR (Pixel-aligned Expressive humAn mesh Recovery), which predicts EHM-s parameters to recover a more expressive human model. 
    \item   We introduce a two-stage training scheme leveraging differentiable rendering. This provides dense pixel-level feedback, significantly reducing misalignment.
    \item  We propose a part-level human parameter annotation strategy, enabling \ourname~to generalize across  diverse cropping scenarios (from close-up faces to whole bodies). The processed data and pipeline will be released to the research community.
    \item  We demonstrate that a single ViT-B backbone is sufficient for high-fidelity body mesh recovery. Our method achieves real-time performance (>100 FPS), paving the way for practical deployment in resource-constrained applications.
\end{itemize}

\section{Related Work}

\textbf{Human Pose and Shape Estimation.} Human pose estimation from images is a well-studied problem with numerous applications \cite{tome2019xr,zhu2024champ,lin2023one,xiang2019monocular,mehta2017monocular,pavlakos2019expressive}. Early optimization-based approaches (e.g., SMPLify-X\cite{pavlakos2019expressive}) estimate body parameters from a single image through iterative fitting, but these methods are often time-consuming. Human Mesh Recovery (HMR) \cite{kanazawa2018end}  alleviates this limitation by directly regressing SMPL \cite{SMPL:2015} parameters with a CNN, significantly reducing inference time. This idea has inspired a series of follow-up methods such as SPIN \cite{kolotouros2019spin} and PARE \cite{kocabas2021pare}, which further improved the accuracy of human pose estimation and extended parameter regression to models like SMPLX \cite{SMPL-X:2019} and MANO \cite{MANO:SIGGRAPHASIA:2017}. More recently, the emergence of ViT-based  methods has led to notable gains in estimation accuracy.  Following previous advances \cite{lin2023one, baradel2024multi,goel2023humans,xia2025hsmr}, we also adopt a transformer-based neural network for EHM-s regression.

\textbf{Human Appearance  Reconstruction.} 
Traditional human reconstruction methods \cite{danvevcek2022emoca,chen2022mobrecon,yuan2022glamr,saito2019pifu} have primarily focused on mesh reconstruction, covering various parts such as the body, face, and hands. 
CAR \cite{liao2023car}, SITH \cite{ho2024sith}, and CanonicalFusion \cite{shin2025canonicalfusion} respectively reconstruct animatable human bodies from single-view and multi-view images. ICON \cite{xiu2022icon} and ECON \cite{xiu2023econ} reconstruct clothed humans through implicit and explicit normal-fusion approaches.
In recent years, the emergence of neural radiance fields  (NeRF) and 3D Gaussian splatting (3DGS) \cite{kerbl20233d} has inspired numerous efforts \cite{weng2022humannerf,zhao2023havatar,yuan2024gavatar,hu2024gauhuman,lei2024gart,liu2024animatable,xu2025seqavatar} to combine human appearance with template models to achieve more realistic 3D reconstructions. Methods such as GART \cite{lei2024gart}, GaussianAvatar \cite{hu2024gaussianavatar} and ExAvatar \cite{moon2024expressive} are typically trained per individual ID, and thus lack generalization ability. More recently, approaches including Human-LRM \cite{weng2024template}, Human-Splat \cite{pan2024humansplat}, LHM \cite{qiu2025lhm}, and GUAVA \cite{zhang2025guava} have focused on human appearance modeling with generalization capability. We refer to this class of methods as Neural Renderers, which can provide pixel-level supervisory signals.
Recently, this field remains highly active, with numerous outstanding works \cite{wu2025localdygs,sun2024aios,patel2025camerahmr,stathopoulos2024score,wang2025prompthmr,shen2025adhmr,shin2025canonicalfusion,wuswift4d}, focusing on expressive human pose estimation and 3D human or scene reconstruction.

\section{Method}
\label{sec:method}

\vspace{1em}
\begin{figure}[t]
    \centering
    \includegraphics[width= 0.7\linewidth]{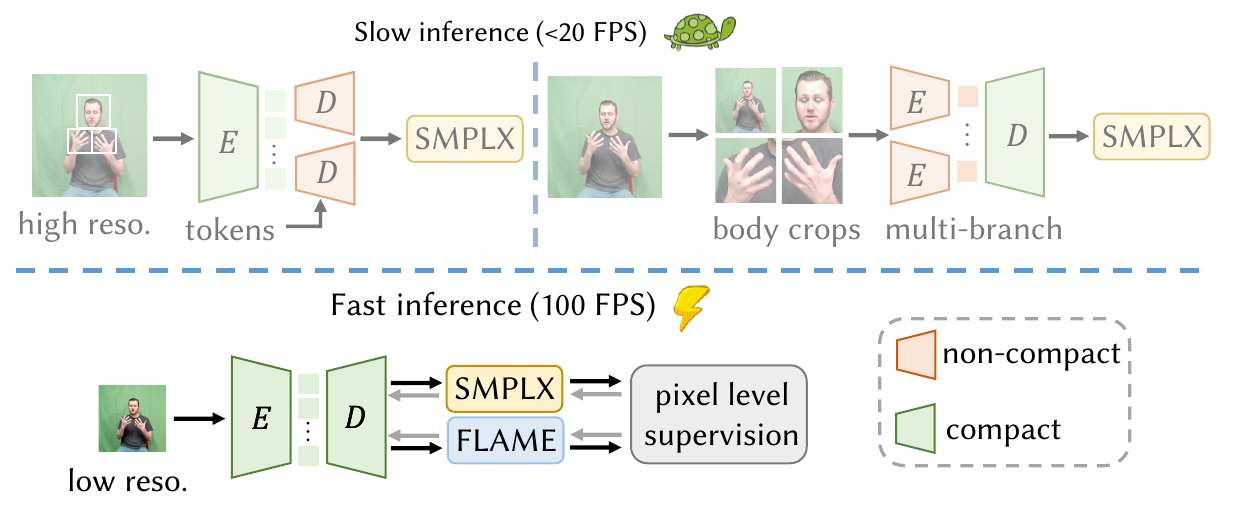}
    \caption{ \textbf{Expressive human mesh recovery  method comparison.} Existing  SMPLX-based methods mainly adopt the two architectures shown on the top. In contrast, we introduce pixel-level supervision without increasing model complexity, enabling fast inference with high-quality body and face reconstruction. This is not trivial, and details are provided below. }
    \label{figs: Expressive HMR Method comparison}
    \vspace{-1em}
\end{figure}

We introduce \ourname~to overcome three fundamental limitations of current SMPLX-based approaches:
slow inference speed, limited fine-grained pose localization accuracy, and insufficient robustness to diverse cropping conditions.  Prior SMPLX–based methods typically rely on high-resolution inputs or multi-branch networks dedicated to hands and face, which leads to slow inference and prevents real-time deployment. Moreover, they primarily emphasize body pose accuracy, often overlooking regions with richer details such as the face and hands. To overcome these issues, we adopt a single, streamlined ViT as the backbone to ensure fast inference. To maintain its performance, we reprocessed a large-scale human dataset and designed a two-stage training strategy to further enhance the model's ability to capture fine-grained details.

\subsection{Preliminary}
As shown in Fig.\ref{fig: methods comp}, existing human mesh recovery methods can be broadly categorized into three types according to the scope of predicted human components. Type1 methods recover body-only parameters, Type2 methods jointly estimate body and hand poses, and Type3 methods further incorporate facial expression and shape modeling.
As models predict an increasing number of human components, the resulting human representations become more expressive and applicable to a wider range of scenarios, accompanied by increased modeling complexity and task difficulty.
\textbf{Our method builds upon SMPLX and FLAME by introducing an additional scale parameter $s$, forming EHM-s }(see supplementary for details), and thus falls into the Type-3 category. Accordingly, our comparisons focus on methods within the same category that aim to recover full-body, hand, and facial parameters within a unified framework.

\subsection{Model Architectures}
\label{sec:Architecture}

\begin{figure}[t]
    \centering
    \includegraphics[width= 0.8\linewidth]{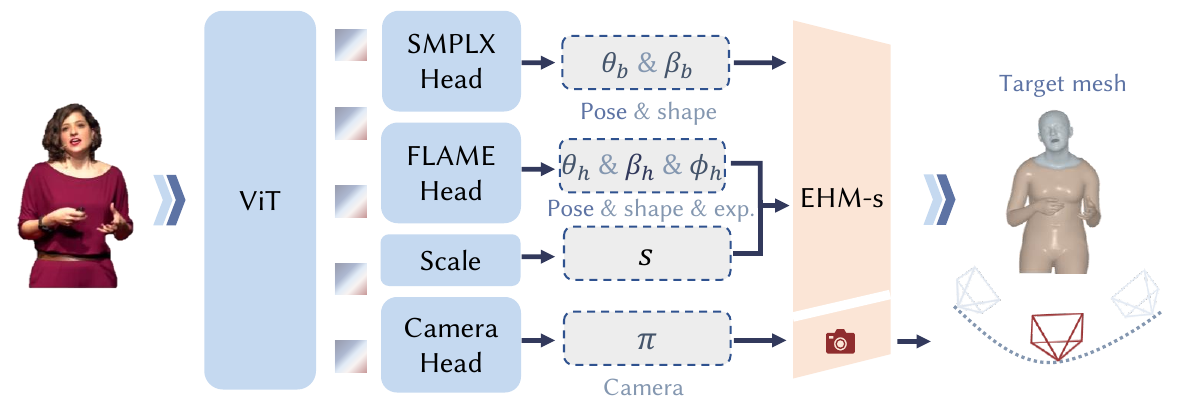}
    \caption{\textbf{ Overview of  \ourname.} \ourname~ adopts a unified ViT backbone (includes encoder and decoder) to jointly regress SMPLX body parameters and FLAME-consistent head parameters. A scale parameter $s$ is introduced to account for head size variations, enabling robust modeling across both children and adults while maintaining efficient inference.
    }
    \vspace{-1em}
    \label{figs:pipelins}
\end{figure}

As shown in Fig. \ref{figs: Expressive HMR Method comparison}, we observe that prior SMPLX methods often rely on high-resolution inputs \cite{baradel2024multi} or multi-branch networks \cite{yin2025smplest,lin2023one,zhang2023pymaf} to handle facial and hand details. While these designs can improve performance, they increase inference cost and hinder real-time deployment. To ensure efficiency, we adopt a single, streamlined ViT-B backbone \cite{dosovitskiy2020image,xu2022vitpose} to simultaneously predict body, face, and hand parameters from a single image. Additionally, to enhance facial realism, we replace the estimated SMPLX face parameters with FLAME parameters, which provide stronger facial expression modeling. The overall pipeline is illustrated in Fig. \ref{figs:pipelins}. As with previous explorations, a single ViT alone cannot fully capture high-dimensional human parameters, so we introduce a second-stage pixel level refinement, as described in Sec. \ref{sec:refine}.

For the model architecture, we design the facial output as a separate branch predicting FLAME parameters, allowing it to independently infer facial parameters even without body input, which makes the model more robust. The model consists of two main head branches, SMPLX and FLAME, each responsible for a distinct set of parameters:
\textbf{SMPLX}: pose parameters $\theta_b$ and shape parameters $\beta_b$. 
\textbf{FLAME}: pose parameters $\theta_h$, shape parameters $\beta_h$, and expression parameters $\phi_h$. 
The above process can be formulated as follows: given an input image $I$, the human pose model $F_{vit}$ predicts $[\theta_b, \beta_b, \theta_h, \beta_h, \phi_h, s, \pi] = F_{vit}(I).$ 
For the body and hand supervision, the mean squared error (MSE) losses are applied to the SMPLX parameters:
\begin{equation}
\mathcal{L}_{body}=||\theta_b-\theta_b^*||_2^2 + ||\beta_b - \beta_b^*||_2^2,
\end{equation}
where $\theta_b^*$ and $\beta_b^*$ denote the ground-truth SMPLX pose parameters and shape parameters, respectively. We also incorporate 3D and 2D body keypoints as supervision, using the L1 loss:
\begin{equation}
\mathcal{L}_{kp1}=||X_b-X_b^*||_1 + || \pi(X_b) - x_b^*||_1,
\end{equation}
where $X_b^*$ and $x_b^*$ denote the ground-truth 3D and 2D  body keypoints, respectively, and $X_b$ represents the predicted 3D keypoints. $\pi(X_b)$ denotes the projection of the 3D points onto the image plane using camera parameters $\pi$.

For supervision of the FLAME parameters, we adopt an $L_1$ loss:
\begin{equation}
\mathcal{L}_{\text{head}} =
\lVert \theta_h - \theta_h^* \rVert_1 +
\lVert \beta_h - \beta_h^* \rVert_1 +
\lVert \phi_h - \phi_h^* \rVert_1  +
\lVert s - s^* \rVert_1  ,
\end{equation}
where $\theta_h^*$, $\beta_h^*$, $\phi_h^*$ and $s^*$ denote the ground-truth FLAME pose, shape, expression and scale  parameters, respectively.

In addition to parameter supervision, we further impose keypoint-based supervision. To account for variations in head scale across different subjects, we first apply a scale parameter $s \in \mathbb{R}^3$ to the predicted head geometry:
\begin{equation}
V_h = s \cdot F_{\text{FLAME}}(\theta_h, \beta_h, \phi_h), \qquad
X_h = F_{\text{v2k}}(V_h),
\end{equation}
where $V_h$ and $X_h$ denote the predicted head vertices and facial keypoints, respectively. Here, $F_{\text{FLAME}}$ represents the FLAME model function, and $F_{\text{v2k}}$ denotes the vertex-to-keypoint mapping. Finally, the facial keypoint supervision is defined as:
\begin{equation}
\mathcal{L}_{\text{kp2}} = \lVert \pi(X_h) - x_h^* \rVert_1 ,
\end{equation}
where $\pi(\cdot)$ denotes the projection function and $x_h^*$ represents the ground-truth 2D facial keypoints.


\begin{figure}[t]
\centering
  \includegraphics[width=0.7\linewidth, trim= 0cm 0cm 0cm 0cm ]{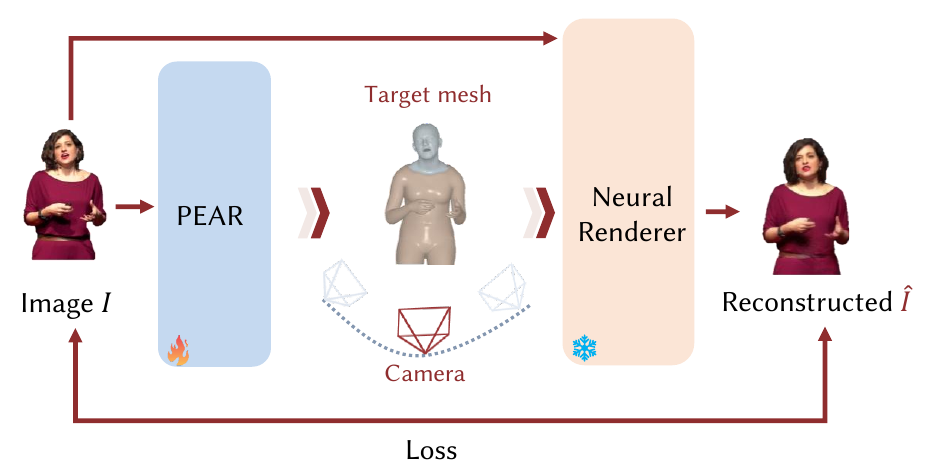}
    \caption{Pixel-aligned training strategy for enhancing \ourname.}
    \label{fig:enhancepear}
\end{figure}

\subsection{Pixel-level Enhancement}
\label{sec:refine}

Although a single ViT-B backbone ensures efficient inference, it is inherently limited in modeling high-dimensional fine-grained details. Furthermore, supervision restricted to body parameters and keypoints frequently introduces local misalignments. Together, these issues compromise the reliability of downstream tasks. To address these issues, we incorporate a pre-trained 3DGS-based neural renderer \cite{zhang2025guava} to provide pixel-level supervision, improving the alignment of predicted meshes with image pixels in detail-rich regions. 
Specifically, as shown in Fig. \ref{fig:enhancepear}, for image data, we randomly select image $I$, and feed them into the human pose estimation model $F_{vit}$ to extract the human body model parameters as follows: 
\begin{equation}
\Phi  = [\theta_b, \beta_b, \theta_h, \beta_h, \phi_h, s, \pi]= F_{vit}(I).
\end{equation}
Subsequently,  the final reconstructed image can be expressed as:
\begin{equation}
\hat I = F_{ren}(F_{ehm}(\Phi),I, \pi).
\end{equation}
Here, $F_{ehm}$ and $F_{ren}$ denote the EHM-s model and the neural renderer. With this formulation, we can seamlessly introduce the photometric loss to provide pixel-level supervision:
\begin{equation}
\mathcal{L}_{photo} = \mathcal{L}_{1}(I,\hat I ) + \mathcal{L}_{lpips}(I,\hat I ).
\end{equation}

As shown in Fig. \ref{fig:enhancepear}, this improvement cannot be achieved without a carefully designed training and supervision strategy. \textbf{It requires a strong prior: the predicted mesh must be roughly aligned with the image for the Gaussian points to be correctly bound during rendering; otherwise, severe coupling between appearance and geometry may occur.}  This motivates our two-stage training strategy (stage 1 for coarse mesh, stage 2 for fine mesh, see Sec. \ref{sec: details}) and the creation of a more accurately annotated dataset to provide reliable supervision. Together, these contributions enable \ourname~ to achieve high-fidelity, real-time full-body reconstruction—a combination rarely achieved by prior methods. Finally, the ablation study is presented in Tab. \ref{tab:ablation}. 
To make this approach practical, we need to generate accurate supervision signals, as detailed below.



\begin{figure}[h]
    \centering
    \vspace{-5pt}
    \includegraphics[width=0.5\linewidth]{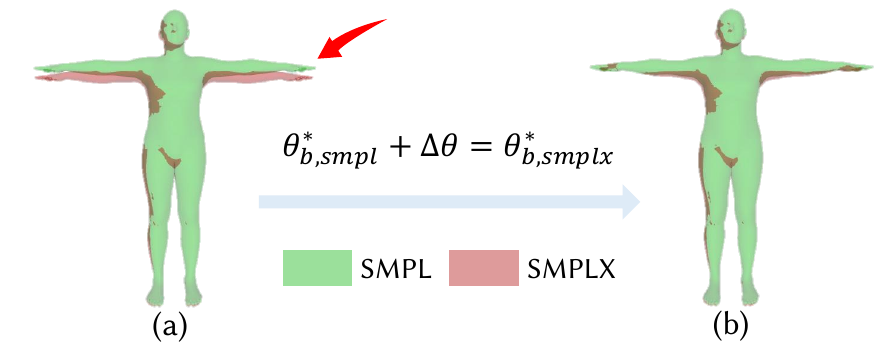}
    \caption{Alignment of SMPL body pose to SMPLX, disregarding the hands and face.}
    \label{fig:smpl2smplx}
    \vspace{-10pt}
\end{figure}

\subsection{Training Data Generation: Part-level pseudo-label refinement strategy }
\label{sec:data_generation}

For learning-based methods, high-quality data supervision is a crucial factor influencing model performance \cite{zhou2023lima}. 
Existing SMPLX–based methods predominantly depend on pseudo ground-truth annotations produced by earlier SMPLX pipelines, which constrains their performance and prevents further improvement. In addition, we find that \textbf{SMPLX–based approaches lag behind SMPL-based methods in body pose estimation accuracy.} Consequently, reusing such SMPLX pipelines for pseudo-label generation \textbf{inevitably propagates their limitations}. To overcome this bottleneck, we introduce a part-level human parameter annotation strategy that independently annotates body, face, and hand components.
This enables the construction of a dataset covering full-body poses, facial expressions, and hand articulations, providing more reliable and comprehensive supervision. 
Moreover, this scheme significantly improves the model’s generalization across different human-part inputs, allowing robust inference on head, upper-body, and full-body images—capabilities that prior methods generally lack, as shown in Fig.~\ref{fig:teaser}.

Specifically, we employ ProHMR~\cite{kolotouros2021prohmr} to estimate SMPL body poses and adapt them to SMPL-X with minor adjustments. As illustrated in Fig.~\ref{fig:smpl2smplx}, directly transferring SMPL body pose parameters to SMPL-X leads to inconsistencies due to intrinsic offsets between the two models. To mitigate this issue, we estimate an offset $\Delta \theta$ by aligning their T-poses, enabling SMPL-X body pose parameters to be directly derived from SMPL-annotated datasets:
\begin{equation}
\theta_{b,\text{SMPL-X}}^{*} = \theta_{b,\text{SMPL}}^{*} + \Delta \theta ,
\end{equation}
where $\theta_{b,\text{SMPL-X}}^{*}$ and $\theta_{b,\text{SMPL}}^{*}$ denote the body pose parameters of SMPL-X and SMPL, respectively.

For the hands and face, we adopt recent SOTA methods, namely HAMER~\cite{pavlakos2024reconstructing} and TEASER~\cite{liu2025teaser}, to obtain coarse initial estimates. Specifically, HAMER is used to estimate SMPL-X hand pose parameters, while TEASER predicts FLAME facial shape $\beta_h^*$, expression $\phi_h^*$, and pose $\theta_h^*$. Subsequently, we use the 2D keypoints estimated by DWPose \cite{yang2023effective} to refine the facial and hand parameters. \textbf{These efforts yield high-quality supervision data, essential for training a pixel-aligned and robust HMR model like our \ourname.} More details can be found in the supplementary material.

\subsection{Implementation Details}
\label{sec: details}



Overall, our method adopts a two-stage training strategy. \textbf{In the first stage}, we train a ViT-based model $F_{vit}$  on large-scale image attribute datasets to estimate the parameters of the EHM-s model. This stage is trained for approximately 200k iterations with a batch size of 40 on 8 NVIDIA A6000 GPUs, taking about 10 days to complete. Afterward, we fine-tune the GUAVA model on full-body human datasets to enhance its generalization ability to images containing complete human body structures.
\textbf{In the second stage}, we leverage pixel-level supervision provided by GUAVA to further refine fine-grained human body details and improve the pixel-wise alignment between the predicted EHM-s human mesh and the input images. This stage is trained for 20k iterations with a batch size of 2, also on 8 NVIDIA A6000 GPUs, and takes about 1 day to finish.

\section{Experiment}
\label{sec:experment}

\begin{table*}[htbp]
\centering
\begin{minipage}{0.4\textwidth}
    \centering
    \caption{Quantitative comparison of human heads. Among them, the LVE metric is the most important evaluation criterion. $^*$ denotes a dedicated facial expression capture method. }
    \label{tab:head_exp}
    \begin{threeparttable}
    \setlength{\tabcolsep}{3pt}
\resizebox{1\textwidth}{!}
{
    \begin{tabular}{l|cc|cccc}
        \toprule
        \multirow{3}{*}{Methods} & \multicolumn{2}{c|}{UBody} & \multicolumn{2}{c}{3DPW} \\
        \cmidrule{2-5}
        & MLE$\downarrow$ & LVE$\downarrow$  & MLE$\downarrow$ & LVE$\downarrow$ \\
     & (\scriptsize$\times 10^{-3}$ m) & (\scriptsize$\times 10^{-5}$ m)  & (\scriptsize$\times 10^{-3}$ m) & (\scriptsize$\times 10^{-4}$ m) \\
        \midrule
        SMIRK$^*$ \cite{SMIRK} & $2.81$ & $8.02$ & $4.25$ & $2.77$ \\
        TEASER$^*$ \cite{liu2025teaser} & $1.92$ & $4.23$ & $3.95$ & $5.60$ \\
        SMPLest \cite{yin2025smplest} & $8.93$ & $15.6$ & $13.3$ & $9.38$ \\
        Ours   & \textbf{0.72} & \textbf{1.22} & \textbf{3.36} & \textbf{0.99} \\
        \bottomrule
    \end{tabular}
    }
    \end{threeparttable}
\end{minipage}
\hfill
\begin{minipage}{0.55\textwidth}
    \centering
    \caption{Quantitative comparison of human hands (PA-PVE) on the UBody (OSX)   and EHF test dataset. $^1$ Tested on publicly released checkpoints. }
    \label{tab: hand_exp}
    \begin{threeparttable}
    \setlength{\tabcolsep}{3pt}
\resizebox{0.9\textwidth}{!}
{
    \begin{tabular}{l|c|c|cc}
        \toprule
        Method & Backbone  & Reso. & EHF$\downarrow$ & UBody$\downarrow$ \\
        \midrule
    \multirow{1}{*}{PIXIE} \cite{PIXIE:2021}  & RN50 & crop & 11.1 & 12.2 \\
    \multirow{1}{*}{Hand4Whole} \cite{Moon_2022_CVPRW_Hand4Whole} & RN50 &crop & 10.8 & 8.9 \\
    \midrule
    \multirow{1}{*}{OSX} \cite{lin2023one}& ViT-L/16 & $256\times192$ & 15.9 & 10.8 \\
    \multirow{1}{*}{SMPLer-X} \cite{cai2023smpler}  & ViT-L/16 & $256\times192$ & 15.0 & 10.3 \\
    \multirow{1}{*}{Multi-HMR}$^1$ \cite{baradel2024multi}  & ViT-B/14 & $896\times896$ & 16.4 & 11.2 \\
    \multirow{1}{*}{Ours} & ViT-B/16 & $256\times192$ & \textbf{12.8} & \textbf{9.8} \\
        \bottomrule
    \end{tabular}
    }
    \end{threeparttable}
\end{minipage}
\end{table*}

\begin{table*}[t] 
    \centering
    \caption{ Quantitative comparison of human body pose. For fairness, all the baselines considered in this work are open-source methods that are capable of simultaneously estimating body, face, and hand parameters. We report PCK @0.05 and @0.1 for the 2D datasets (COCO, LSP-Extended, PoseTrack) and MPJPE, PA-MPJPE and MVE for the 3D datasets (3DPW, AGORA).   $^1$ Multi-HMR always detects additional persons in the PoseTrack dataset; therefore, we do not report its results. $^2$ test on L40S.}
    \label{tab:body_exp}
    \begin{threeparttable}
    \resizebox{\textwidth}{!}{ 
    \begin{tabular}{l|c|c|c|cc|cc|cc|cc|c}
        \toprule
        \multirow{2}{*}{Methods} & \multirow{2}{*}{Model} & \multirow{2}{*}{Predict parts} & \multirow{2}{*}{FPS $^2$} & \multicolumn{2}{c|}{COCO} & \multicolumn{2}{c|}{LSP-extend} & \multicolumn{2}{c|}{PoseTrack} & \multicolumn{2}{c|}{3DPW} & \multicolumn{1}{c}{AGORA} \\
        \cmidrule{5-13}
         & &&  & @0.05$\uparrow$ & @0.1$\uparrow$ & @0.05$\uparrow$ & @0.1$\uparrow$ & @0.05$\uparrow$ & @0.1$\uparrow$ & MPJPE$\downarrow$ & PA-MPJPE$\downarrow$ & Body MVE$\downarrow$   \\
        \midrule
        OSX \cite{lin2023one}& SMPLX & Body,face,hand & 20 & 0.70 & 0.87 & 0.42 & 0.73 & 0.82 & 0.90 & 74.7 & \textbf{45.1} & 80.2 \\
        Multi-HMR$^1$\cite{baradel2024multi}  & SMPLX  & Body,face,hand & 10 & 0.65 & 0.82 & 0.43 & 0.70 & - & - & 78.0  &  45.9 & 59.6    \\
        SMPLest-X \cite{yin2025smplest}  & SMPLX  & Body,face,hand & 20 & 0.71 & 0.91 & 0.40 & 0.74 & 0.82 & 0.91  & 74.8 & 45.8  & 63.5 \\
        Ours    & EHM-s  & Body,face,hand & \textbf{100} & \textbf{0.81} & \textbf{0.94} & \textbf{0.55} & \textbf{0.81} & \textbf{0.87} & \textbf{0.97}  & \textbf{71.3}   & 45.3  & \textbf{59.3} \\
        \bottomrule
    \end{tabular}
    }
    \end{threeparttable}
\end{table*}


In this section, we conduct both qualitative and quantitative evaluations of our human reconstruction and rendering framework. First, compared with previous SMPLX-based approaches  \cite{zhang2022pymaf,lin2023one,yin2025smplest,baradel2024multi}, we demonstrate that \ourname~not only exhibits strong generalization ability, enabling accurate human model estimation under diverse and complex environments, but also achieves more precise pose estimation in fine-grained body regions with improved pixel-level alignment (Sec.~\ref{subsec: accuracy}). Second, we present ablation studies to validate the effectiveness of our approach (Sec. \ref{subsec: ablation}). Finally, we further show the superior performance of our method in rendering-driven downstream applications (Sec. \ref{subsec: downstram}), highlighting its robustness and accuracy in human pose recovery.

\subsection{Experimental Setup}
\label{subsec:setup}
\textbf{Training Dataset.}  
Our training data consists of two parts. The first part, referred to as \textbf{Part 1}, comprising the following datasets: Human3.6M \cite{h36m_pami}, MPI-INF-3DHP \cite{mono-3dhp2017}, COCO \cite{lin2014microsoft}, MPII \cite{andriluka20142d}, InstaVariety \cite{kanazawa2019learning}, and AVA \cite{gu2018ava}. The second part, referred to as \textbf{Part 2}, includes Ego-Exo4D \cite{grauman2024ego}, Ego-humans \cite{khirodkar2023ego}, SA1B \cite{kirillov2023segment}, Harmony4D \cite{khirodkar2024harmony4d}, and AI Challenger \cite{wu2017ai}. The data annotation process is described in Sec. \ref{sec:data_generation}.

\textbf{Baseline.} We report results on benchmarks commonly used for comparison with a wide range of prior methods. Since our approach integrates both SMPLX and FLAME models, we evaluate against SMPLX-based methods for body and hand reconstruction, and against dedicated FLAME-based methods for face reconstruction. For fairness, all the baselines considered in this work are open-source methods that are capable of simultaneously estimating body, face, and hand parameters. We exclude methods that estimate only the body, or the body and hands, since jointly estimating SMPL-X body, face, and hand parameters poses a greater challenge.


\subsection{Model Evaluations}
\label{subsec: accuracy}
\textbf{Facial expression evaluation metrics.} 
For 3D head mesh modeling, we assess reconstruction accuracy using key metrics such as lip vertex error (LVE) \cite{richard2021meshtalk} and mean vertex error (MVE), which measure the deviation of mouth and overall facial vertices from the tracking results. 
Since no dedicated benchmark exists, we follow the fitting strategy proposed in GUAVA to generate test samples from the 3DPW \cite{vonMarcard2018} and UBody \cite{zhang2025guava} test splits. 
Quantitative comparisons with state-of-the-art learning-based methods, including TEASER, SMIRK, and SMPLest-X (Tab.~\ref{tab:head_exp}), show that our method captures high-quality facial expressions without face cropping, achieving performance comparable to or better than these approaches. In particular, when compared with SMPLX–based methods such as SMPLest-X and OSX, our approach exhibits pronounced advantages in facial expression capture, as illustrated in Fig.\ref{fig: head comp}.

\begin{figure}
   \centering
   \includegraphics[width=\textwidth, trim=0cm 0cm 0cm 0cm ]{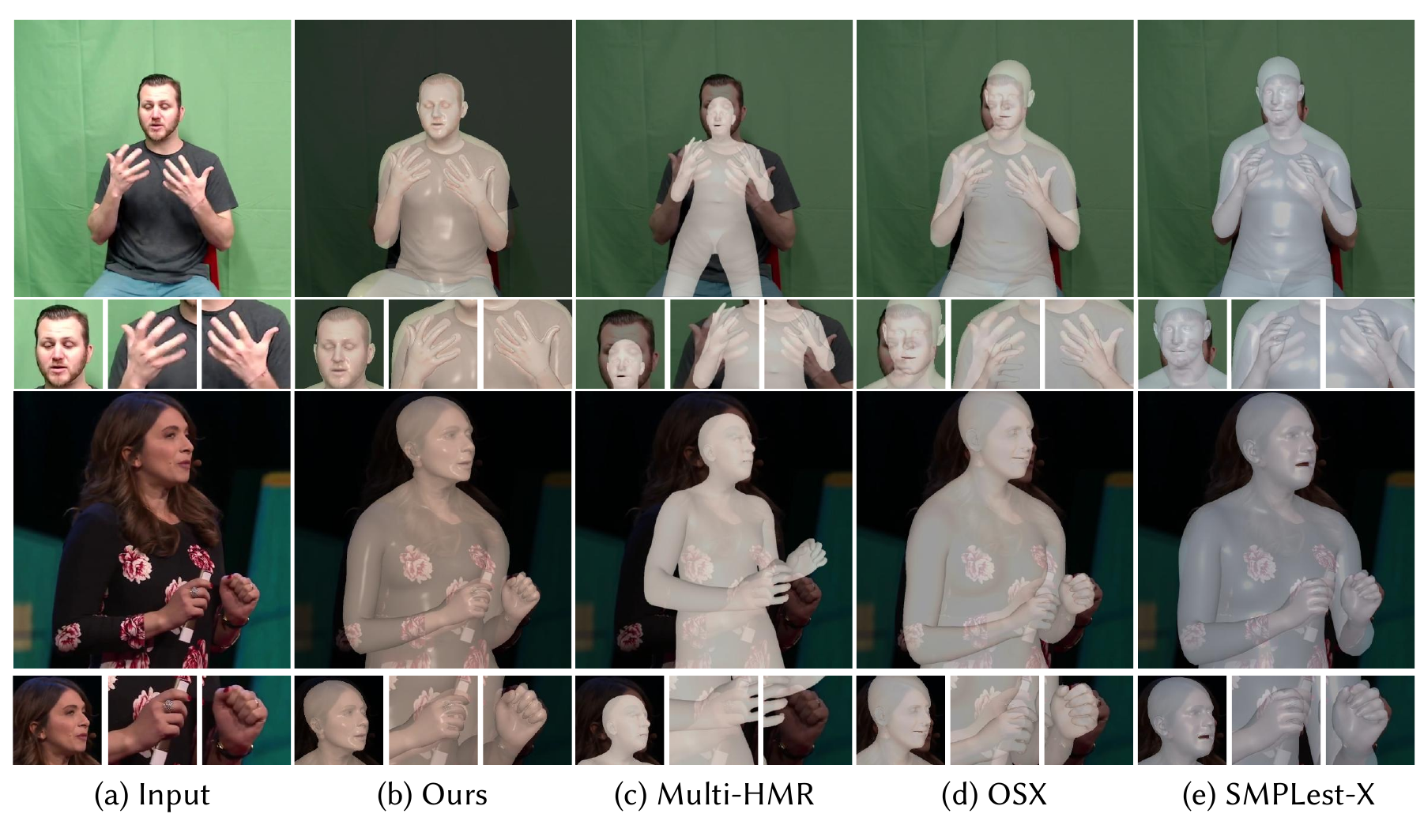}
   \caption{Hand pose prediction comparison. (Results on the Ubody-intra dataset).}
   \vspace{-1em}
   \label{fig: hand comp}
\end{figure}

\begin{figure}
   \centering
   \includegraphics[width=0.8\textwidth]{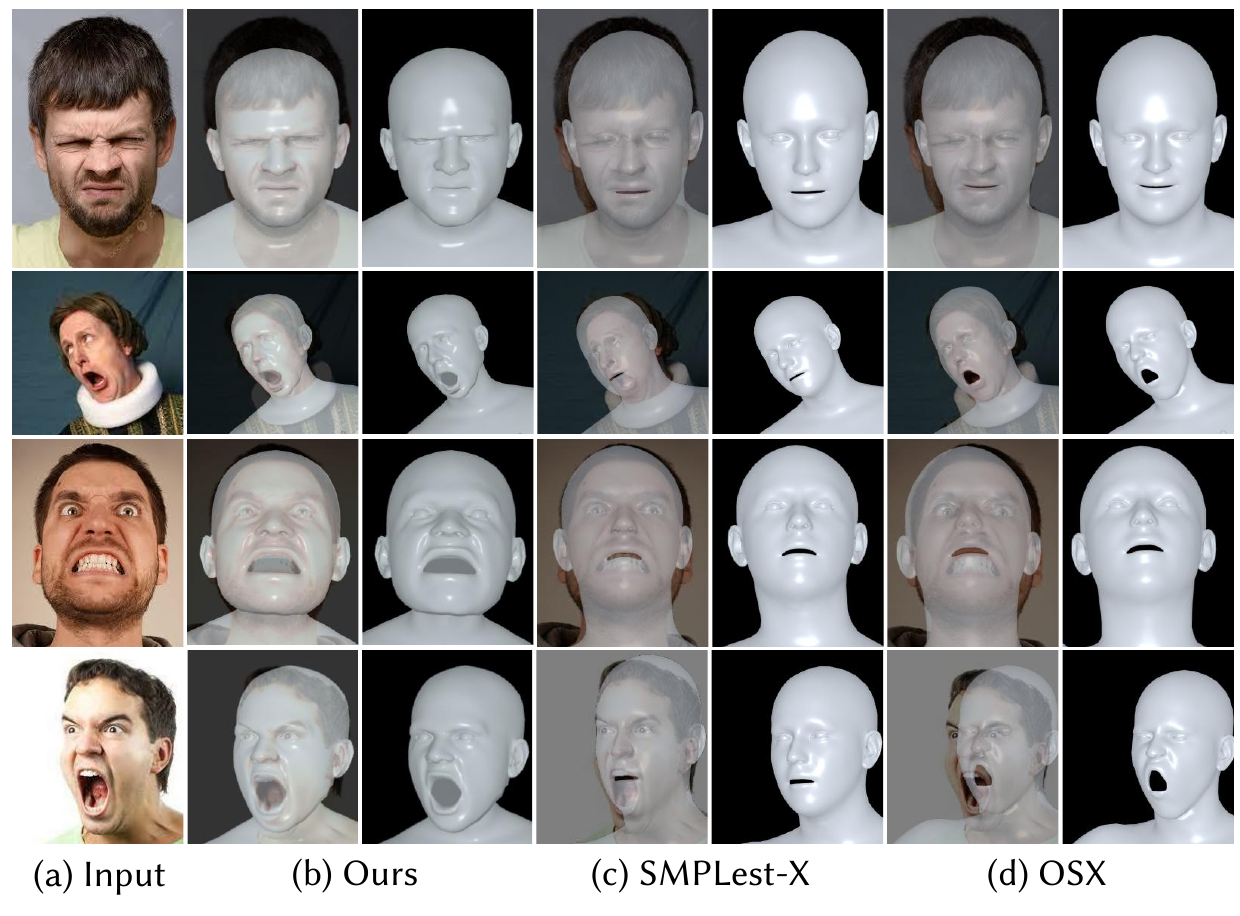}
   \caption{Facial expression capture comparison. Qualitative comparison of facial expression reconstruction between our method and state-of-the-art approaches, including SMPLest and OSX. Our method better preserves fine-grained and expressive facial details. (Examples from publicly available Internet images.)
   }
   \label{fig: head comp}
\end{figure}


\begin{figure}[h]
   \centering
   \vspace{-3em}
   \includegraphics[width=0.8\textwidth]{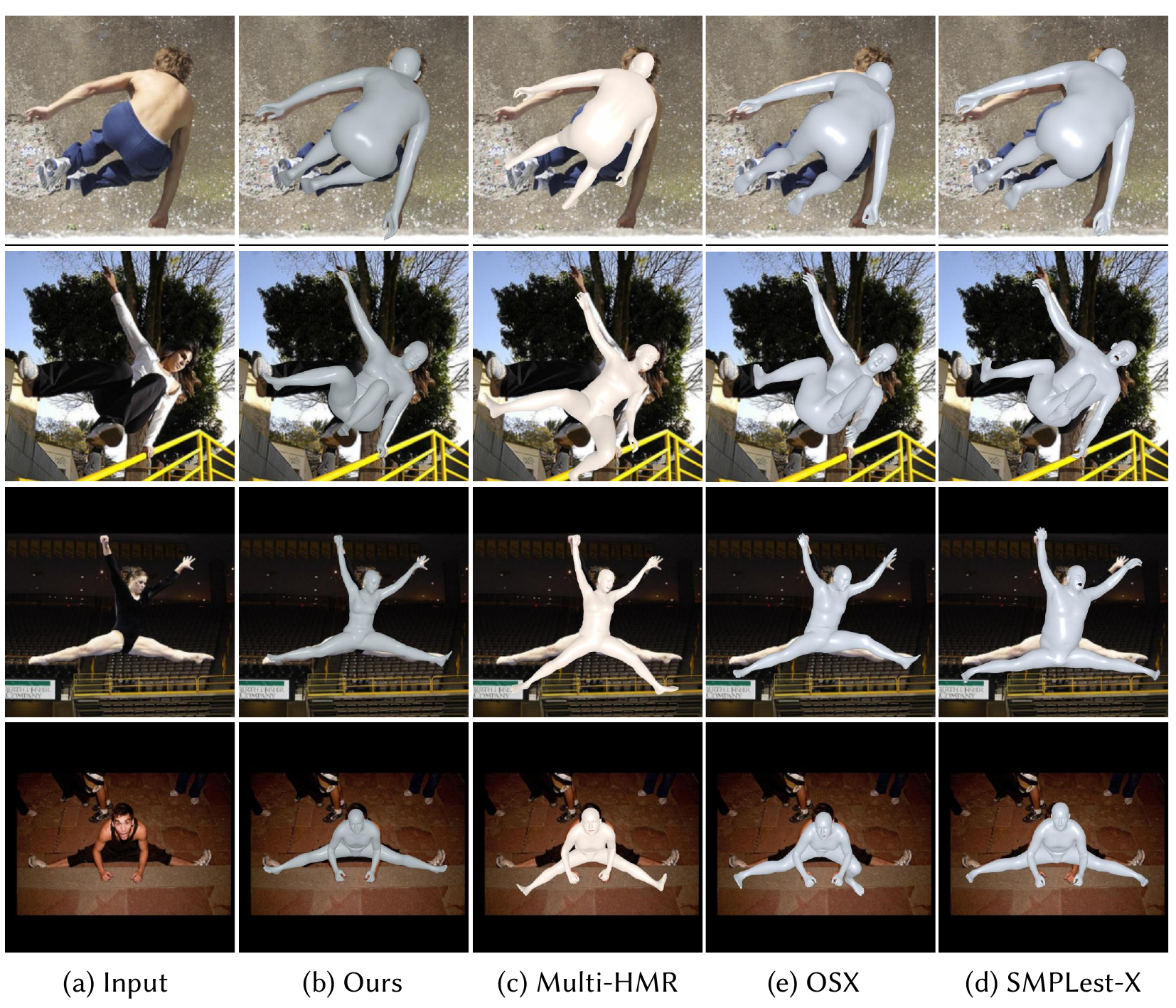}
   \caption{Body pose prediction comparison.  We compare \ourname~ with several smplx-based sota approaches. \ourname~  not only predicts body poses more accurately, but also achieves higher alignment at the pixel level. (Results on the LSP-extend dataset). }
   \label{fig: body comp}
\end{figure}

\begin{table*}[t] 
    \centering
    \caption{\textbf{Training dataset ablation study.} The datasets used in Part~1 and Part~2 are described in Sec.~\ref{subsec:setup}. Incorporating Part~2 further improves overall body accuracy; however, this improvement comes at the cost of reduced hand accuracy.}
    \label{tab:training dataset abla}
    \begin{threeparttable}
    \resizebox{\linewidth}{!}{ 
    \begin{tabular}{l|c|cc|cc|cc|c|c}
        \toprule
        \multirow{2}{*}{Dataset} &   \multirow{2}{*}{Images/Frames} & \multicolumn{2}{c|}{COCO} & \multicolumn{2}{c|}{LSP-extend} & \multicolumn{2}{c|}{PoseTrack} & \multicolumn{1}{c|}{EHF} & \multicolumn{1}{c}{Ubody-intra}\\
        \cmidrule{3-10}
         &  & @0.05$\uparrow$ & @0.1$\uparrow$ & @0.05$\uparrow$ & @0.1$\uparrow$ & @0.05$\uparrow$ & @0.1$\uparrow$ & Hand(P-PVE) $\downarrow$ & Hand(P-PVE) $\downarrow$  \\
        \midrule
        Part1  & 3 M+ & 0.79 & \textbf{0.94} & 0.52 & 0.80 & \textbf{0.87} & \textbf{0.97} & 13.3 & \textbf{9.5} \\
        Part1 + Part2  & 6 M+  & \textbf{0.81} & \textbf{0.94} & \textbf{0.55} & \textbf{0.81} & \textbf{0.87} & \textbf{0.97} & 13.3 & 12.8 \\
        \bottomrule
    \end{tabular}
    }
    \end{threeparttable}
\end{table*}

\textbf{Hand pose evaluation metrics.} 
Following prior works, we evaluate the Procrustes Alignment per-vertex error (PA-PVE) metric for hands on the EHF and UBody-intra \cite{lin2023one} datasets, as shown in Tab. \ref{tab: hand_exp}. Among the baselines, PIXIE \cite{PIXIE:2021} and Hand4Whole \cite{Moon_2022_CVPRW_Hand4Whole} require additional cropping of the hand region, while Multi-HMR \cite{baradel2024multi} uses an input resolution of $896 \times 896$, which results in lower efficiency. In contrast, our method only requires an input resolution of $256 \times 256$ and achieves comparable performance without any extra processing. As shown in Fig. \ref{fig: hand comp}, we compare our method with state-of-the-art approaches for hand pose prediction, demonstrating more accurate hand pose capture.

\textbf{Body pose evaluation metrics.}
We conduct extensive evaluations of body pose estimation accuracy across multiple datasets and a diverse set of metrics. We first assess 2D image alignment of the generated human poses by reporting the Percentage of Correct Keypoints (PCK) of reprojected keypoints under different thresholds, as shown in Tab.~\ref{tab:body_exp}. Our approach significantly outperforms other SMPLX–based methods, indicating that \ourname~ achieves more accurate body pose estimation even on challenging datasets such as COCO and LSP-Extended. 
We additionally evaluate 3D pose accuracy on the 3DPW and AGORA datasets using the Mean  Procrustes-Aligned  per-joint error  (MPJPE) and Mean Vertex Error (MVE). Our method consistently surpasses current sota approaches on both benchmarks.

\textbf{Real-time inference and animation.}
Our method enables fully end-to-end inference of SMPLX and FLAME parameters, eliminating the need for any preprocessing steps.  Our model achieves real-time inference across a wide range of GPU platforms. Specifically, it runs at approximately 100 FPS on both the L40S and RTX 4090, and around 80 FPS on the A100 (40GB and 80GB), with all results reported using FP32 precision.

\begin{table}[h]
\centering
\caption{EHM parameter evaluation and animation time (seconds per frame) on different machine configurations (unit: s, Float32 precision).}
\resizebox{0.7\textwidth}{!}{\begin{tabular}{lccccccc}
\hline
  Operation  & L40S & A100  & RTX 5090 & RTX 4090 & RTX 3090 \\
\hline
EHM-s inference  & 0.009 & 0.013& 0.010 & 0.011 & 0.015 \\
Animation & 0.014 & 0.013& 0.016  & 0.019 & 0.020 \\
Total  & 0.023  & 0.026 & 0.026 & 0.030 & 0.035 \\
\hline
\end{tabular}
}
\vspace{-1em}
\label{table:infer time}
\end{table}

\begin{table}[h]
    \centering

    \caption{Ablation study results (in meters) on 4 different datasets. MLE ($\times10^{-4}$) and LVE ($\times10^{-5}$) measure facial expression accuracy; PA-PVE measures hand vertex accuracy.  } 
    \setlength{\tabcolsep}{2pt}
    \label{tab:ablation}
    \begin{threeparttable}
    \resizebox{0.55\linewidth}{!}{
    \begin{tabular}{lcccccc}
    \toprule
    \multirow{2}{*}{Methods} 
        & \multicolumn{2}{c}{UBody} 
        & \multicolumn{2}{c}{3DPW} 
        & EHF & UBody-intra \\
    \cmidrule(lr){2-3}\cmidrule(lr){4-5}
     & MLE$\downarrow$ & LVE$\downarrow$ 
     & MLE$\downarrow$ & LVE$\downarrow$  
     & PA-PVE$\downarrow$ & PA-PVE$\downarrow$  \\
    \midrule
     w/o $L_{photo}$  & 7.92 & 1.43 & 3.43 & 1.24 & 13.3 & 12.8  \\
     w $L_{photo}$  &  \textbf{7.19}&\textbf{1.22} & \textbf{3.36}  & \textbf{0.99} & \textbf{12.8} & \textbf{9.8}\\
    \bottomrule
    \end{tabular}}
    \end{threeparttable}
\end{table}

\begin{figure}[h]
   \centering
   \includegraphics[width=0.8\textwidth, trim=0.5cm 14cm -0.8cm 0cm]{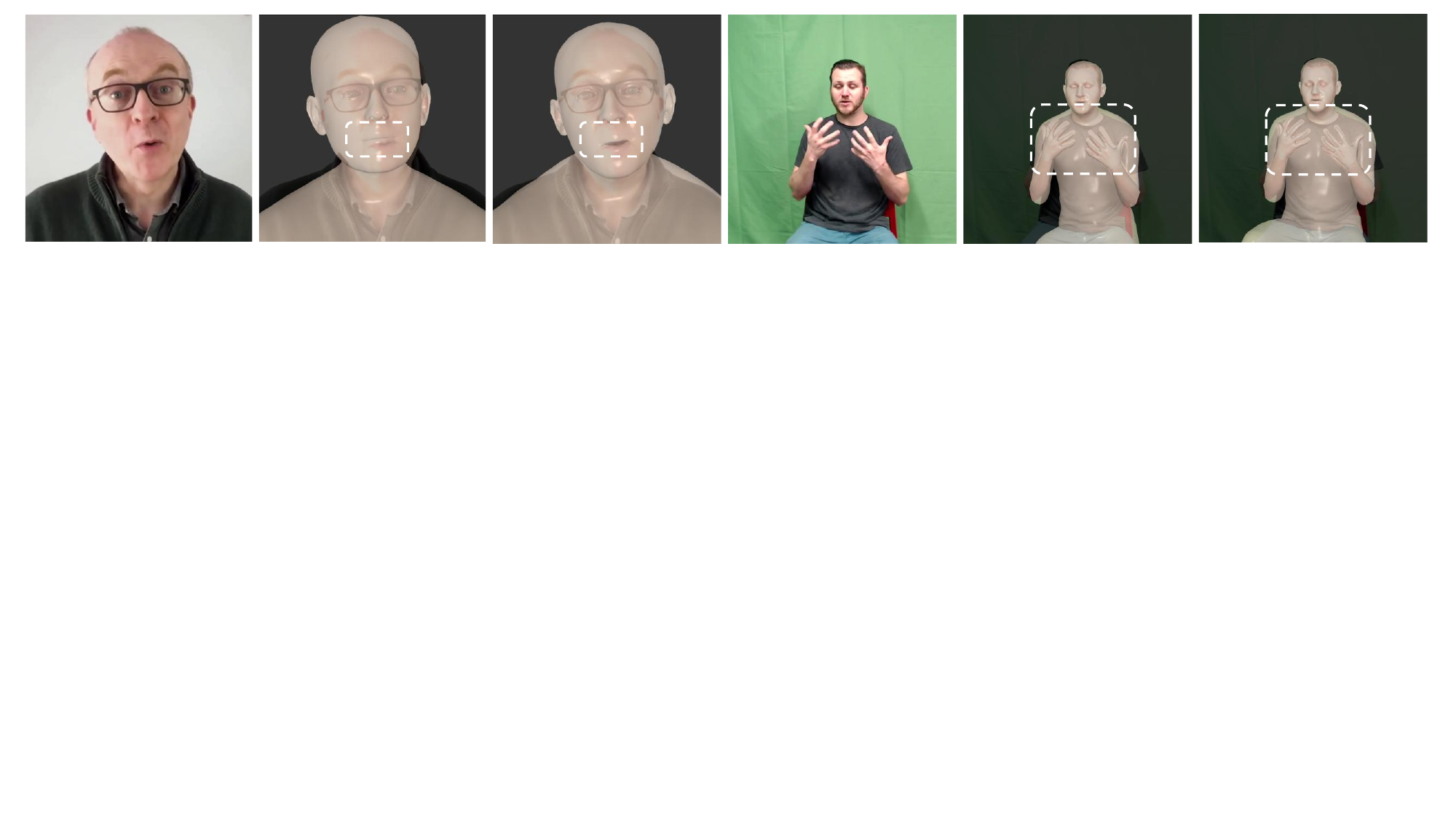}
   \caption{  Ablation on two-stage training. Joint training improves fine-grained body part alignment (left: w/o photometric loss, right: w  photometric loss). }
   \label{fig: l_photo abla}
   \vspace{-3mm}
\end{figure}

\subsection{Ablation Study}
\label{subsec: ablation}
\textbf{Pixel-level supervision ablation.} We conduct a comprehensive ablation study on the proposed photometric loss, as shown in Tab.~\ref{tab:ablation} and Fig. \ref{fig: l_photo abla}. For facial expression accuracy, we follow SMIRK \cite{SMIRK} and adopt the Mean Vertex Error (MVE) and Lip Vertex Error (LVE) as evaluation metrics. For hand pose accuracy, consistent with previous SMPLX–based methods, we quantify performance using the PA-PVE. Experimental results across multiple datasets demonstrate that the proposed photometric loss effectively and consistently enhances the model’s ability to capture fine-grained human details, particularly in expressive regions such as the face and hands, while preserving overall body pose accuracy.

\textbf{Training dataset ablation.}  
As shown in Tab. \ref{tab:training dataset abla}, we observe that incorporating additional training data does improve overall body accuracy; however, it also shifts the model’s attention toward the limbs, leading to degraded modeling capability for the hands. This phenomenon may help explain why current SMPLX–based methods generally struggle to surpass SMPL or SKEL in limb accuracy under challenging poses.

\subsection{Downstream Applications}
\label{subsec: downstram}

Since \ourname~ enables real-time capture of body motion, hand poses, and facial expressions, it is well suited for a wide range of downstream applications, including real-time holographic communication, immersive VR/AR experiences, virtual avatar animation, and human–computer interaction. To this end, we have specifically created several examples, as shown on \textcolor{orange}{Poject Page}.

\section{Discussion and Limitation}
\label{Sec:Discussion}

As shown in Fig. 2, we categorize human body models into three types:  body-only models (e.g., SMPL, SKEL), body–hand models (e.g., SMPLH), and full models incorporating body, hands, and face (e.g., SMPLX, MHR, and EHM-s). Our method targets full-body human mesh recovery with integrated modeling of body, hands, and facial expressions, corresponding to the Type3 (body, hand and face) category defined above. While this setting enables more expressive and holistic human representations, it also introduces increased modeling complexity and optimization difficulty.

We acknowledge that, due to the reduced output space, Type1 (body-only) and Type2 (body and hand) approaches may achieve higher accuracy on body pose estimation in certain aspects, as they are designed to recover fewer human components under more constrained settings. In contrast, Type3 methods trade off some pose accuracy for enhanced expressiveness and modeling capacity, which is essential for applications requiring detailed hand and facial modeling.
To ensure fair and meaningful evaluation, our primary comparisons are conducted with other Type3 methods that share the same recovery targets. For completeness, we additionally include representative comparisons with Type1 and Type2 approaches in the supplementary material.
Despite these advantages, our method still faces limitations when handling extremely blocking and complex interactions, which remain open directions for future research.

\section{Conclusion}
\label{sec: conclusion}

In this paper, we present \ourname, the first human mesh recovery framework that simultaneously regresses both SMPLX and FLAME parameters, addressing three key limitations of prior approaches. 
First, SMPLX exhibits limited capacity in modeling facial details, making it inadequate for representing diverse expressions. 
Second, previous HMR methods often fail to achieve pixel-level alignment in the image plane, leading to misaligned human meshes and constraining their applicability in high-precision downstream tasks. 
Third, existing SMPLX–based methods suffer from slow inference, which severely limits their applicability in real-time scenarios. 
To overcome these challenges, \ourname~ takes a single image as input and jointly estimates SMPLX body and FLAME facial parameters, achieving more expressive human meshes at 100 FPS. Overall, our method achieves pixel-level accuracy in HMR and significantly improves the fidelity and utility of human mesh representations in downstream real-time applications.

\begin{figure*}
   \centering
   \includegraphics[width=\textwidth, trim=0cm 3.5cm 0cm 0cm]{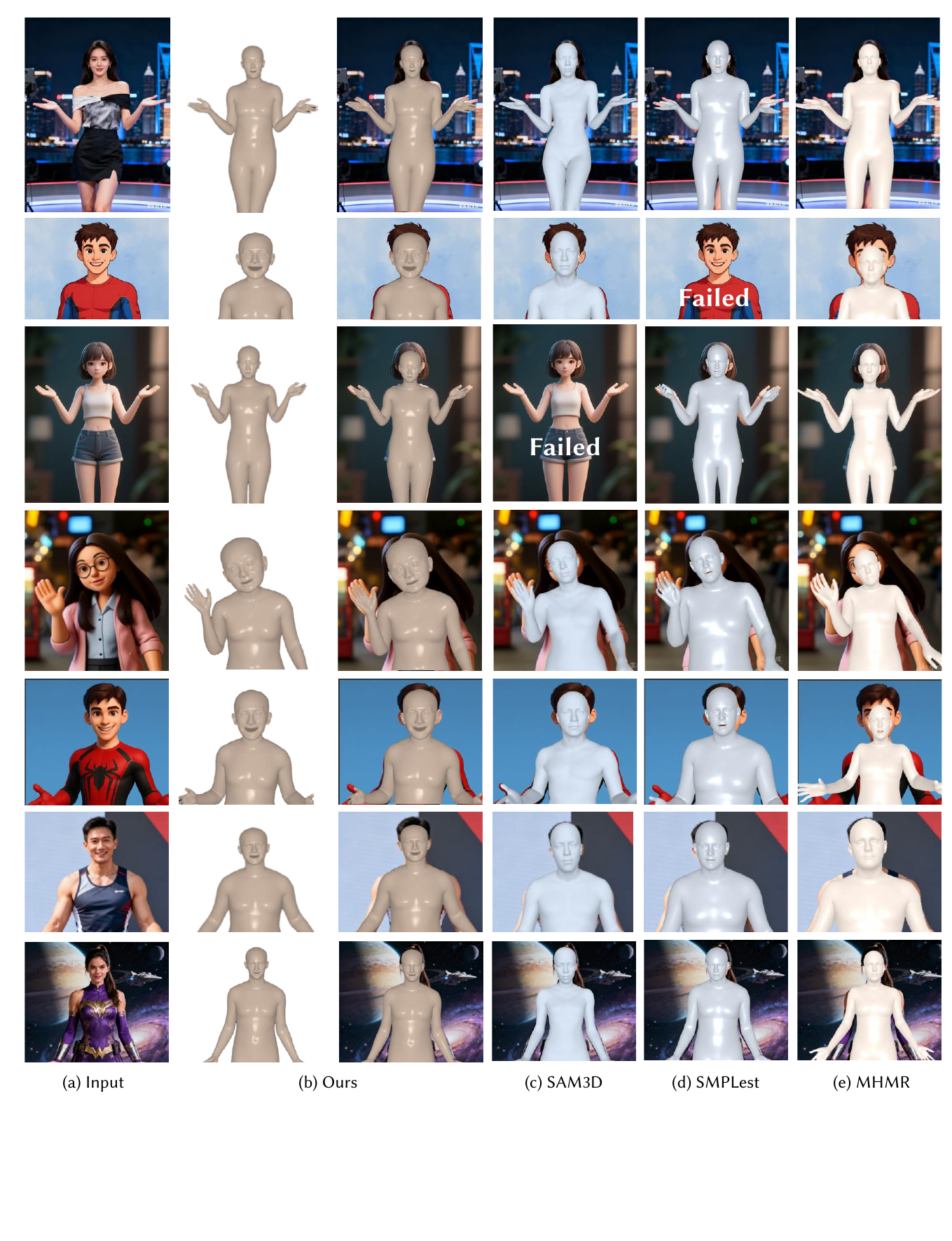}
   \caption{Our model generalizes to cartoon characters, accurately estimating facial parameters even with exaggerated head proportions, which is not supported by prior SMPLX-, or MHR-based methods. (Qualitative results on AI-generated images.)}
   \label{fig: body comp}
\end{figure*}
\begin{figure}
   \centering
   \includegraphics[width=0.8\linewidth, trim=0cm 1cm 0cm 0cm]{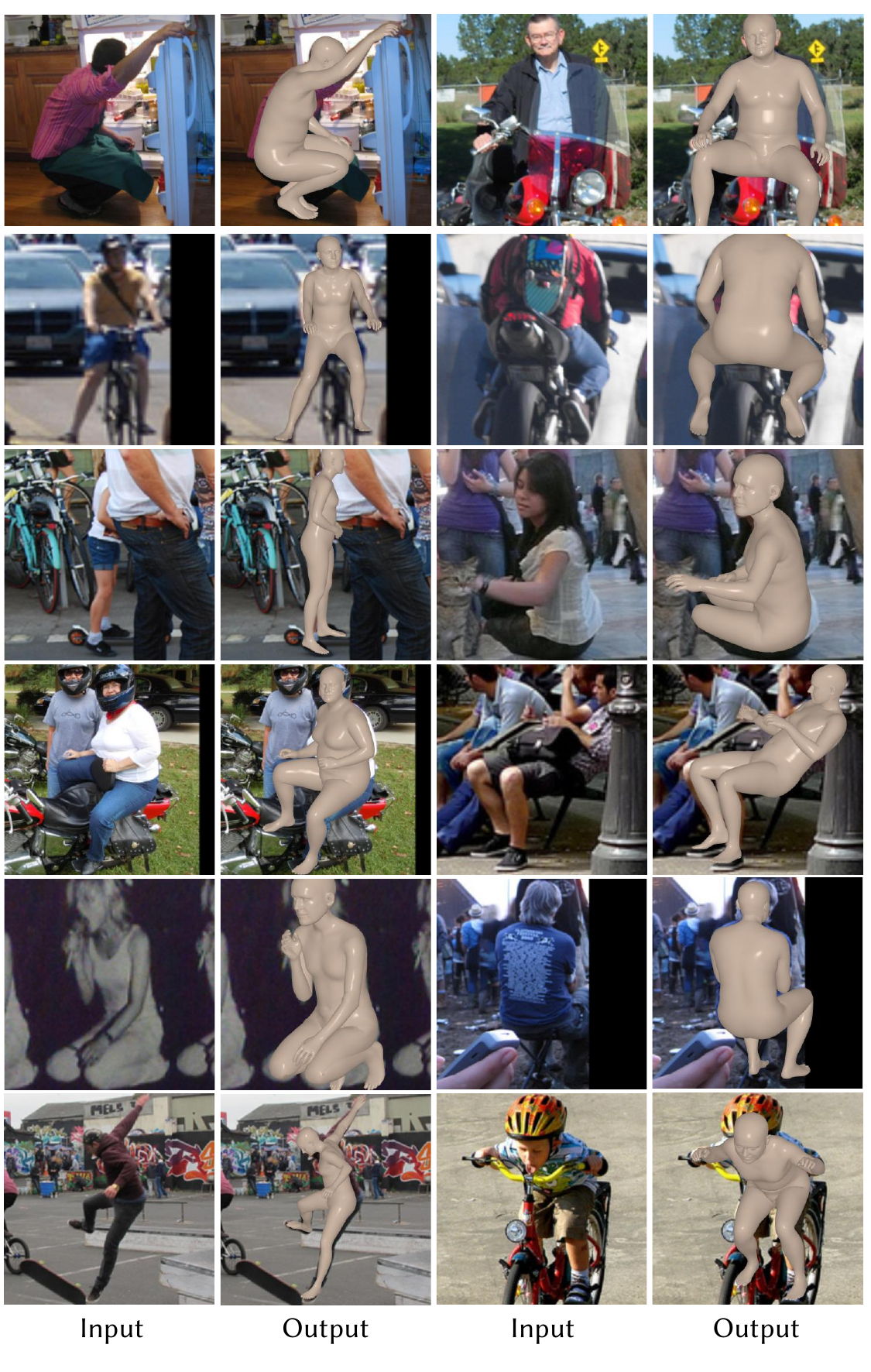}
   \caption{Visualization results on the COCO-val dataset. Our method performs robustly across diverse scenarios, including occlusions, motion blur, and children.
   }
   \label{fig: coco pose}
\end{figure}

\begin{figure}
   \centering
   \includegraphics[width=0.9\linewidth, trim=0cm 1cm 0cm 0cm]{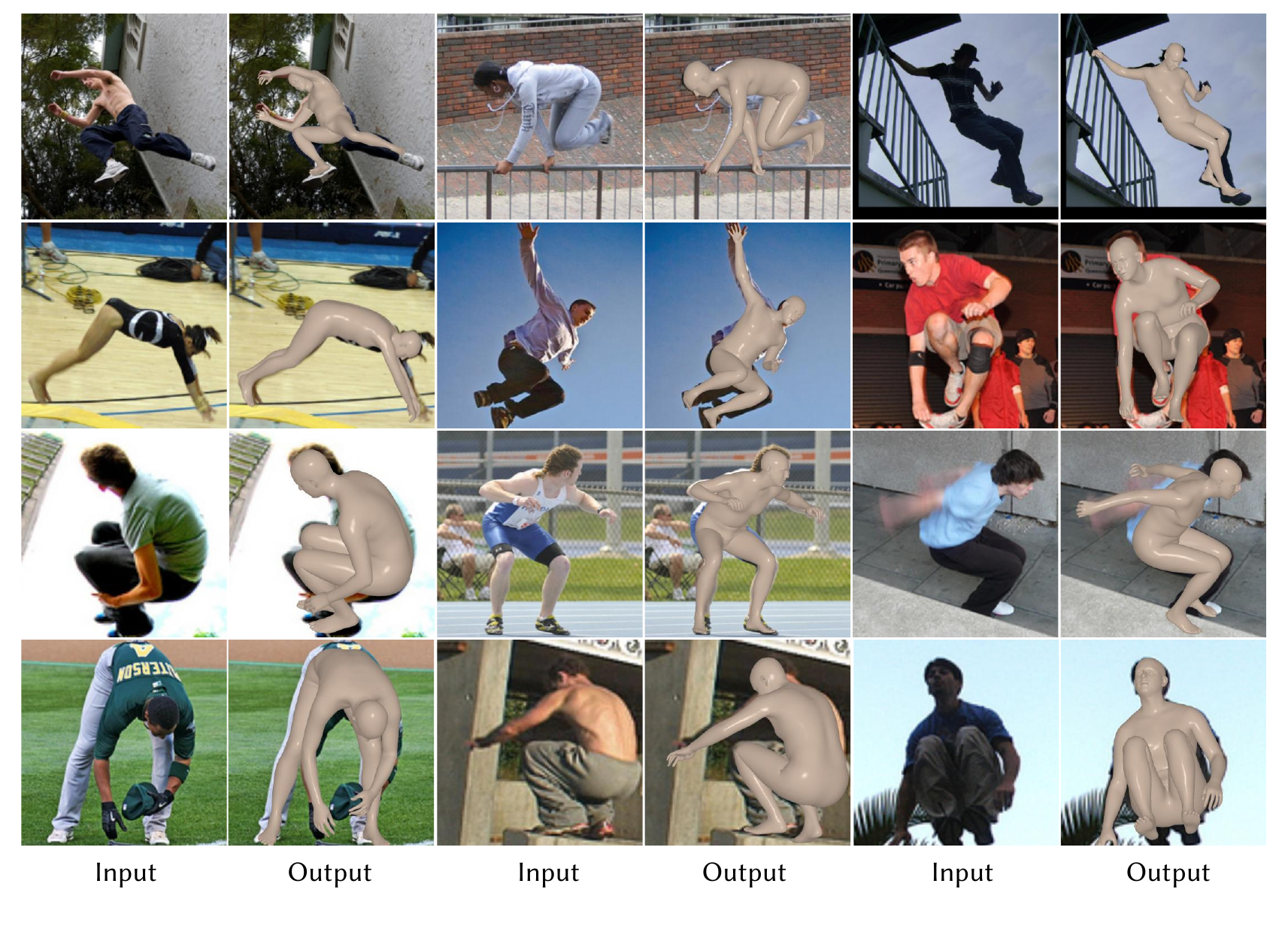}
   \caption{We further visualize the performance of our model on the LSP-Extend dataset, which poses significant challenges for HMR.
   }
   \label{fig: complex pose}
\end{figure}

\begin{figure}
   \centering
   \includegraphics[width=0.9\linewidth, trim=0cm 0.5cm 0cm 0cm]{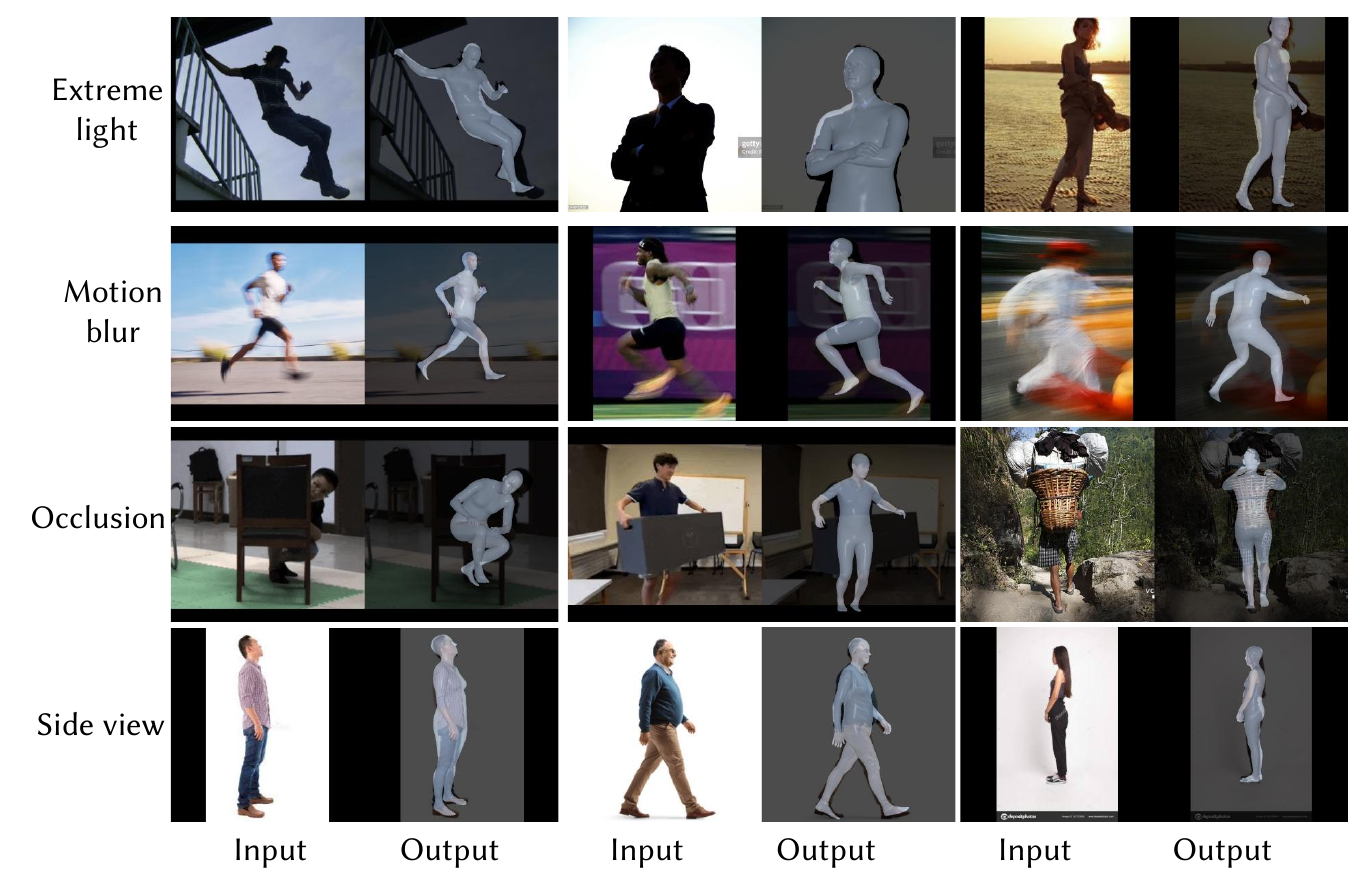}
   \caption{We visualize our model’s performance on challenging and extreme cases, including extreme lighting, motion blur, occlusion, and side views.
   }
   \label{fig: extreme case}
\end{figure}

\begin{figure}
   \centering
   \includegraphics[width=0.8\linewidth, trim=0cm 6cm 0cm 0cm]{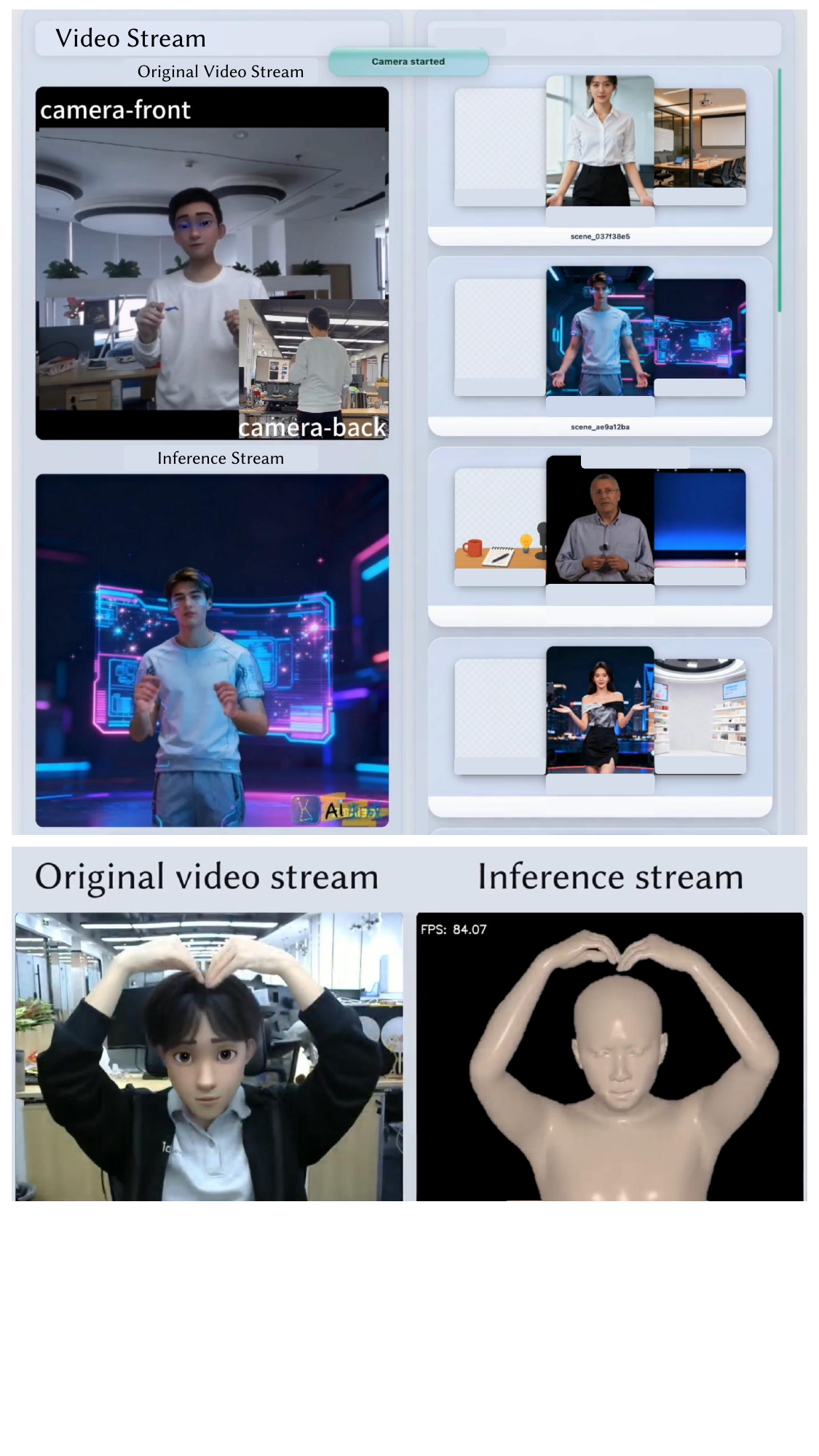}
   \caption{We develop a human-friendly interface for real-time avatar animation (top) and human mesh recovery (bottom). Please find more results in the supplementary video.
   }
   \label{fig: ui}
\end{figure}

\newpage


\bibliographystyle{ACM-Reference-Format}
\bibliography{main}

\newpage

\appendix

\newpage

\appendix

\vspace{1em}
\begin{figure}[t]
    \centering
    \includegraphics[width= 0.7\linewidth]{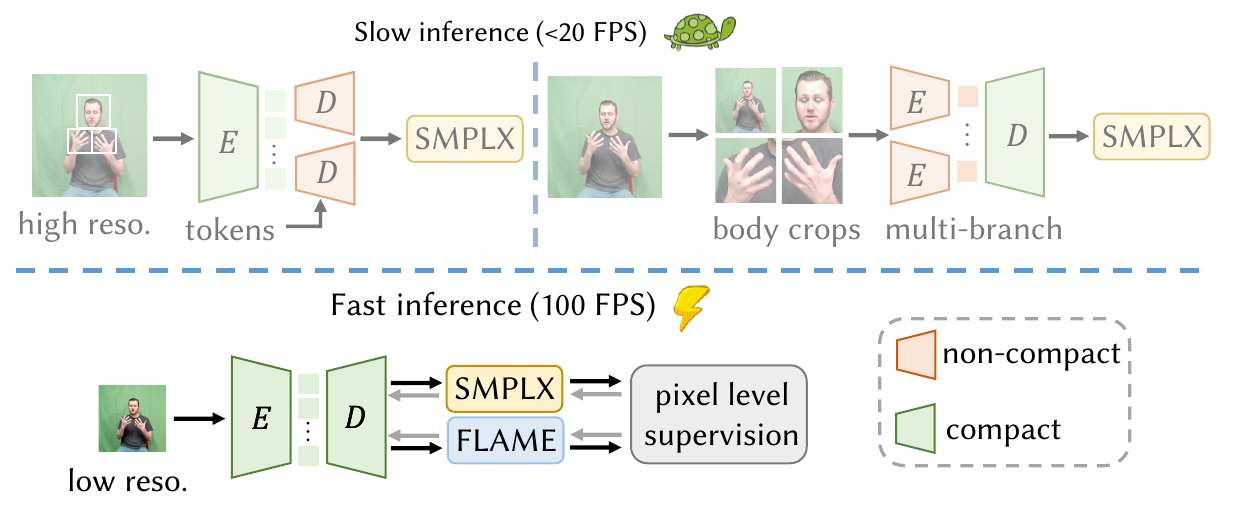}
    \caption{ \textbf{Expressive human mesh recovery  method comparison.} Existing  SMPLX-based methods mainly adopt the two architectures shown on the top. In contrast, we introduce pixel-level supervision without increasing model complexity, enabling fast inference with high-quality body and face reconstruction. This is not trivial, and details are provided below. }
    \label{figs: Expressive HMR Method comparison}
    \vspace{-1em}
\end{figure}

\section{Method details}

\subsection{Our Motivation} 

As shown in Fig. \ref{figs: Expressive HMR Method comparison}, existing expressive human mesh recovery methods mainly follow two paradigms to model high-dimensional human parameters. Although effective, these approaches rely on complex multi-branch architectures and incur heavy computational costs, which hinder real-time deployment. To address this, we remove the multi-branch design and adopt a single lightweight ViT-B backbone that directly predicts high-dimensional SMPLX and FLAME parameters from low-resolution inputs in real time.

During this process, we identify two key challenges of this strategy. \textbf{First}, existing data preprocessing pipelines jointly handle face, hand, and body parameters, leading to limited annotation accuracy and reduced data diversity (e.g., a lack of close-up facial images). We therefore redesign the data processing pipeline to provide higher-quality supervision. \textbf{Second}, we observe that a simple ViT struggles to capture fine-grained details from low-resolution inputs, motivating us to introduce pixel-level supervision for further refinement. As a result, we achieve an efficient and accurate model that enables real-time estimation of both SMPLX and FLAME parameters, supporting downstream applications. We are the first to explore a simple ViT-based architecture for extracting full-dimensional human body parameters and to validate the feasibility of this approach, achieving fast inference at 100 FPS and aiming to provide a fast and accurate interface for downstream applications.

\subsection{Kinematic Consistency of EHM-s}

\textbf{Here, we describe in detail how the EHM-s model adheres to plausible human body structure.}  EHM-s is constructed by combining SMPL-X with a scaled FLAME head. The alignment procedure is performed as follows:

1. We first obtain a refined, expression-aware head mesh by passing the predicted shape and expression parameters through FLAME.  

2. Next, we generate the SMPL-X template using the corresponding SMPL-X shape and expression parameters.  

3. In the canonical space of SMPL-X, we perform joint-based registration to align the head mesh, replace the corresponding head region, and obtain the SMPL-X template with the integrated head.  

4. Finally, the updated template from the previous step is processed through LBS to produce the final fused mesh.

The corresponding pseudo-code is presented in Tab. \ref{alg:flame_smplx_fusion}

\begin{figure}
   \centering
   \includegraphics[width=0.7\textwidth, trim=0cm 1cm 0cm 0cm]{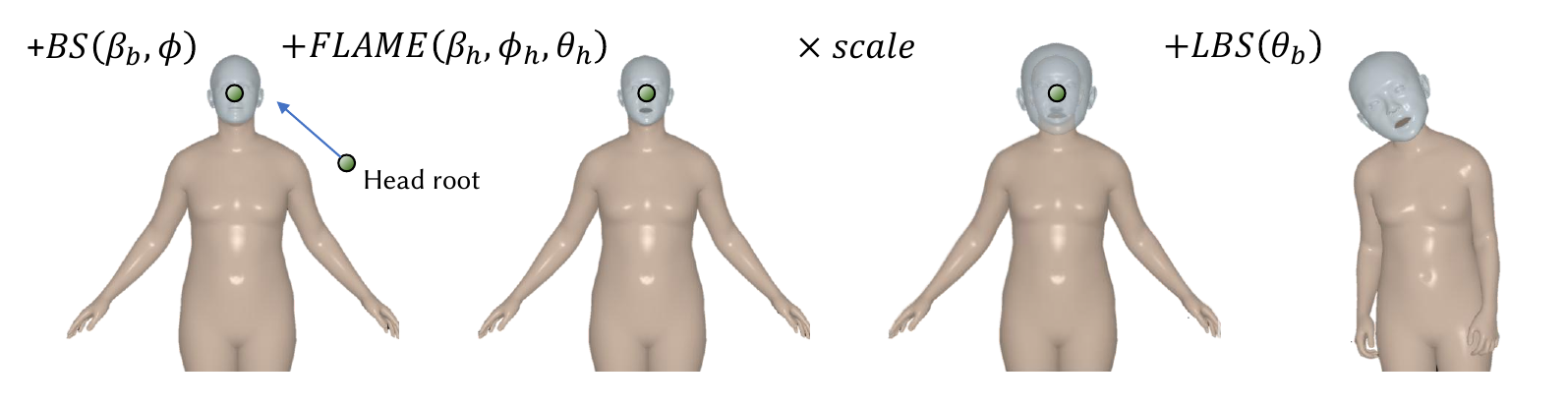}
   \caption{ The EHM-s forward pipeline.}
   \label{fig: ehm forward}
\end{figure}

\begin{figure}
   \centering
   \includegraphics[width=0.6\textwidth, trim=0cm 0.5cm 0cm 0cm]{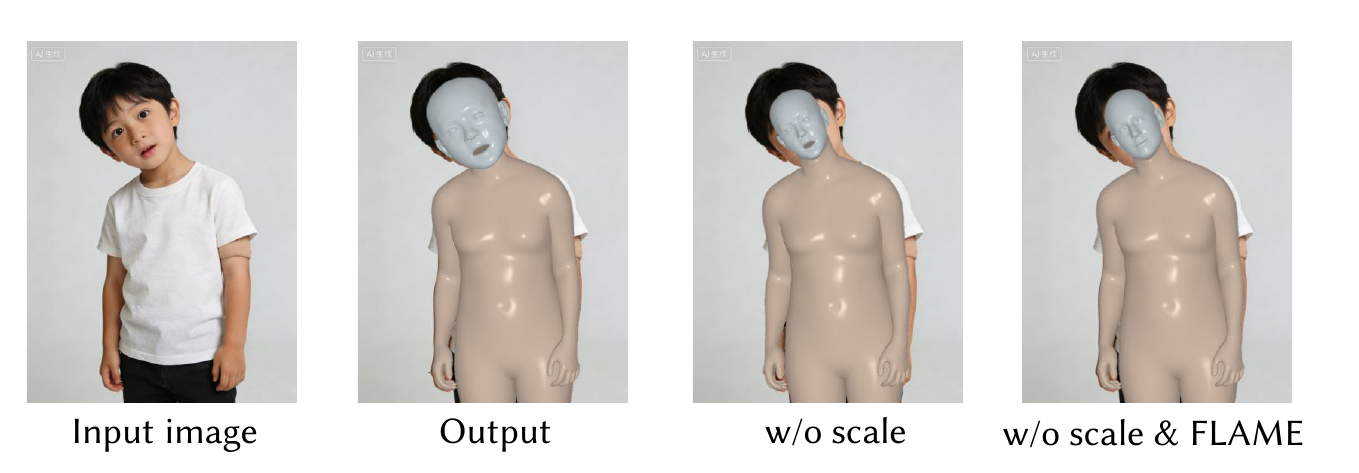}
   \caption{ Effect of different EHM-s parameters on the resulting mesh.}
   \label{fig: ehm forward}
\end{figure}

\begin{algorithm}[t]
\caption{FLAME--SMPL-X Head Fusion Pipeline}
\label{alg:flame_smplx_fusion}
\begin{algorithmic}

\State \textbf{Input:} Head pose parameters, body pose parameters, FLAME head template, head scale $s$, SMPL-X body template
\State \textbf{Output:} Fused full-body mesh vertices

\State \textbf{Step 1: Compute FLAME head mesh via LBS}
\State $(\mathbf{V}_h, \mathbf{J}_h) \leftarrow \mathrm{LBS}(\text{head\_pose}, \text{template\_head\_vertices})$
\Comment{Global pose removed}

\State \textbf{Step 2: Align FLAME head to SMPL-X body}
\State $\mathbf{c}_h \leftarrow \mathrm{Mean}(\mathbf{J}_h[3{:}5])$
\State $\mathbf{c}_b \leftarrow \mathrm{Mean}(\mathbf{J}_b[23{:}25])$
\State $\hat{\mathbf{V}}_h \leftarrow \mathbf{V}_h *s - \mathbf{c}_h + \mathbf{c}_b$

\State \textbf{Step 3: Replace SMPL-X head region}
\State $\mathbf{V}_b[\text{smplx2flame\_indices}] \leftarrow \hat{\mathbf{V}}_h$ 

\State \textbf{Step 4: Final LBS for full-body mesh}
\State $(\mathbf{V}, \mathbf{J}) \leftarrow \mathrm{LBS}(\text{body\_pose}, \text{global\_pose}, \mathbf{V}_b)$

\end{algorithmic}
\end{algorithm}

\begin{figure*}[h]
   \centering
   \includegraphics[width=0.9\textwidth, trim=0cm 1cm 0cm 0cm]{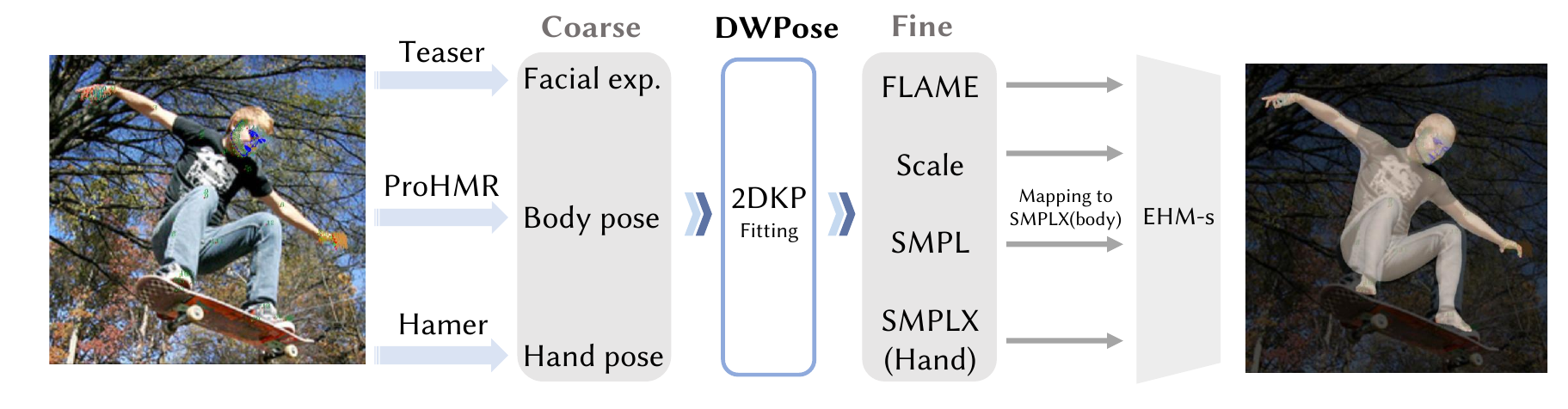}
   \caption{Our Pseudo-Label Generation Pipeline with Part-Level Refinement.
   }
   \label{fig: data generation}
\end{figure*}

\subsection{EHM-s forward process}

As shown in Fig. \ref{fig: ehm forward}, we first start from the SMPL-X basis and apply blend shape deformations using the body shape parameters \( \beta_b \) and expression parameters \( \phi \) to obtain a coarse body template. Next, based on the FLAME basis, we generate accurate and highly expressive facial geometry using the facial shape \( \beta_h \), expression \( \phi_h \), and pose \( \theta_h \) parameters. Since the head root of FLAME is aligned with that of SMPL-X, the head geometry can be replaced and scaled around this shared root to produce a unified and accurate human geometry. Finally, the resulting geometry is posed using the SMPL-X linear blend skinning (LBS) function, yielding the final human mesh.

\section{Training Data Generation: Part-level pseudo-label refinement strategy}

To construct reliable supervision signals for training, we design a novel part-level pseudo-label refinement pipeline, as illustrated in Fig.~\ref{fig: data generation}. Given a single RGB image containing a human subject, we first estimate coarse parametric representations for different body parts using three off-the-shelf models specialized for distinct components: TEASER \cite{liu2025teaser} for facial expression parameters, ProHMR \cite{kolotouros2021prohmr} for full-body pose and shape, and HaMeR \cite{pavlakos2024reconstructing}  for detailed hand pose estimation. These models operate independently and provide initial, part-specific predictions that may suffer from inconsistencies and accumulated errors.

To further improve the accuracy and cross-part consistency of these estimates, we refine all extracted parameters under the supervision of 2D keypoints predicted by the DWPOSE \cite{yang2023effective} model. Specifically, the predicted 2D keypoints serve as geometric constraints to align the projected model joints with image observations, enabling effective correction of pose ambiguities and scale discrepancies.

Through this refinement process, we obtain four sets of optimized parameters, including FLAME parameters for facial geometry and expression, global scale factors, SMPL parameters for the body, and SMPL-X (hand) parameters for articulated hand poses. For body representations, we further convert the refined SMPL parameters into the SMPL-X formulation using a standard parameter mapping, resulting in consistent SMPL-X (body) parameters that are compatible with the hand and face models.

Ultimately, this pipeline produces accurate and coherent SMPL-X and FLAME pseudo-labels, together with reliable scale estimates, which serve as high-quality supervision for subsequent model training. For learning-based methods, a well-designed annotation pipeline is critical.   \textbf{Our modular annotation procedure is thus an indispensable component  for enabling the effective training and performance of our model.}

\begin{figure*}[h]
   \centering
   \includegraphics[width=\textwidth, trim=0cm 1cm 1cm 0cm]{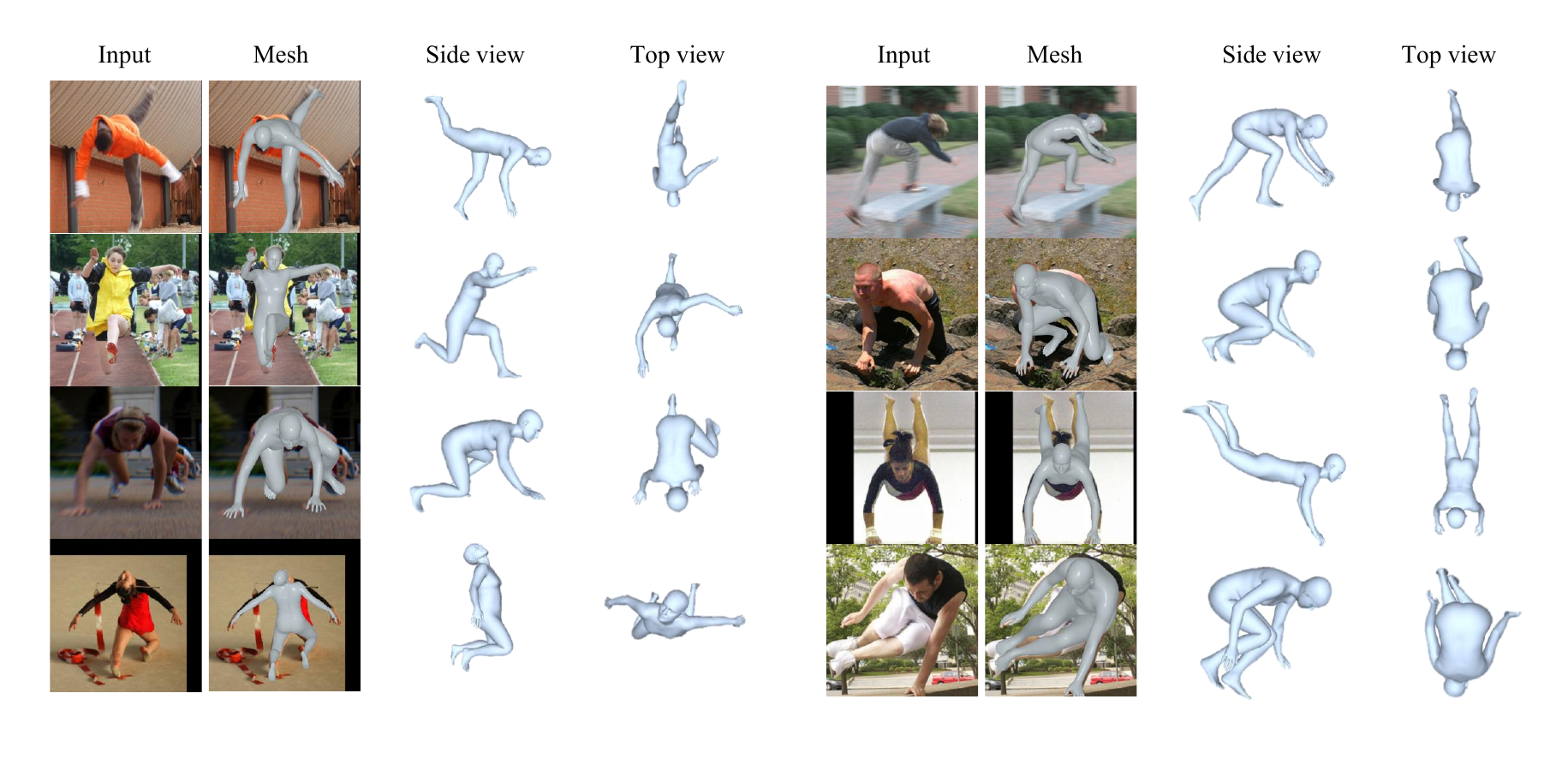}
   \caption{We further visualize the performance of our model on the LSP-Extend dataset, which poses significant challenges for human pose estimation methods.
   }
   \label{fig: complex pose}
\end{figure*}

\subsection{Implementation and training details.}

Our training objective consists of a combination of parameter-level supervision and geometric consistency losses. Specifically, we apply a head parameter loss to supervise fine-grained facial representations, including eye pose, jaw articulation, eyelid motion, expression, and shape parameters, using an L1 loss masked by the availability of FLAME annotations. For full-body modeling, we introduce a body parameter loss that supervises SMPL/SMPL-X parameters, including body pose, hand pose, shape, expression, and scale. Body and hand poses are converted from axis-angle representations to rotation matrices before computing an MSE loss, which improves stability in pose regression. The supervision is selectively applied using validity masks to handle incomplete annotations.

In addition to parameter regression, we incorporate 2D keypoint reprojection loss and 3D keypoint loss to enforce geometric consistency. The 2D loss aligns projected joints with image observations using confidence-weighted L1 distances, while the 3D loss supervises root-relative joint positions to preserve plausible human body structures. The final loss is computed as a weighted sum of all components, ensuring balanced optimization across facial details, articulated body motion, and global geometric alignment.

\subsection{More Results}

We further provide additional qualitative visualizations for analysis, as shown in Fig.~\ref{fig: complex pose}. The results illustrate that our method is capable of producing reliable reconstructions under diverse and complex poses. In particular, the inferred geometry remains stable and accurate when viewed from side and top perspectives, demonstrating consistent spatial reasoning across different viewpoints.

\end{document}